\documentclass{article}

\usepackage{microtype}
\usepackage{graphicx}
\usepackage{booktabs} % for professional tables
\usepackage{times}
\usepackage{epsfig}
\usepackage{amsmath}
\usepackage{amssymb}
\usepackage{booktabs}
\usepackage{caption}
\usepackage[table,xcdraw,dvipsnames]{xcolor}
\usepackage{multirow}
\usepackage{subcaption}
\usepackage{enumitem}
\usepackage{balance}
\usepackage{overpic}

\makeatletter
\newcommand*\bigcdot{\mathpalette\bigcdot@{.5}}
\newcommand*\bigcdot@[2]{\mathbin{\vcenter{\hbox{\scalebox{#2}{$\m@th#1\bullet$}}}}}
\makeatother

\definecolor{c2}{HTML}{FBD9BD}
\definecolor{c3}{HTML}{fe793d}
\definecolor{c4}{HTML}{eedeb0}
\definecolor{c5}{HTML}{00FFFF}
\definecolor{c6}{HTML}{FF00FF}

\definecolor{rouse}{rgb}{0.981,0.961,0.941}

\usepackage[pagebackref=true,breaklinks=true,letterpaper=true,colorlinks,bookmarks=false]{hyperref}

\usepackage[accepted]{icml2026}

\usepackage{amsmath}
\usepackage{amssymb}
\usepackage{mathtools}
\usepackage{amsthm}

\usepackage[capitalize]{cleveref}
\crefname{section}{Sec.}{Secs.}
\Crefname{section}{Section}{Sections}
\Crefname{table}{Table}{Tables}
\theoremstyle{plain}

\theoremstyle{definition}

\theoremstyle{remark}

\usepackage[textsize=tiny]{todonotes}

\icmltitlerunning{Dual-Phase Curriculum Learning for Context-Entangled Segmentation}

\begin{document}

\twocolumn[
  \icmltitle{Refining Context-Entangled Content Segmentation via Curriculum Selection and Anti-Curriculum Promotion
  % Beyond Easy-to-Hard: Dual-Phase Curriculum Learning for Dense Prediction
  % Curriculum Learning for Context-dependent Concept Segmentation
  }

  % It is OKAY to include author information, even for blind submissions: the
  % style file will automatically remove it for you unless you've provided
  % the [accepted] option to the icml2026 package.

  % List of affiliations: The first argument should be a (short) identifier you
  % will use later to specify author affiliations Academic affiliations
  % should list Department, University, City, Region, Country Industry
  % affiliations should list Company, City, Region, Country

  % You can specify symbols, otherwise they are numbered in order. Ideally, you
  % should not use this facility. Affiliations will be numbered in order of
  % appearance and this is the preferred way.
  \icmlsetsymbol{equal}{*}

  \begin{icmlauthorlist}
    \icmlauthor{Chunming He}{yyy}
    \icmlauthor{Rihan Zhang}{yyy}
    \icmlauthor{Fengyang Xiao}{yyy}
    \icmlauthor{Dingming Zhang}{yyy}
    \icmlauthor{Zhiwen Cao}{comp}
    \icmlauthor{Sina Farsiu}{yyy}
  \end{icmlauthorlist}

  \icmlaffiliation{yyy}{Duke University}
  \icmlaffiliation{comp}{Adobe}
  % \icmlaffiliation{*}{Equal contribution}
  % \icmlaffiliation{sch}{School of ZZZ, Institute of WWW, Location, Country}
% \thanks{The first}
\icmlcorrespondingauthor{Fengyang Xiao}{fengyang.xiao@duke.edu}
  \icmlcorrespondingauthor{Sina Farsiu}{sina.farsiu@duke.edu}

  \icmlkeywords{Machine Learning, ICML}

% {\vspace{0.75em}

% 	% \includegraphics[width=0.99\linewidth]{Figure/Intro.pdf}\vspace{-4mm}   
% 	\captionof{figure}{Results of existing COS methods.
%  }
% 	\label{fig:Intro}
% 	\vspace{-5mm}
% % \end{figure*}
% }

  \vskip 0.3in
]

% this must go after the closing bracket ] following \twocolumn[ ...

% This command actually creates the footnote in the first column listing the
% affiliations and the copyright notice. The command takes one argument, which
% is text to display at the start of the footnote. The \icmlEqualContribution
% command is standard text for equal contribution. Remove it (just {}) if you
% do not need this facility.

% Use ONE of the following lines. DO NOT remove the command.
% If you have no special notice, KEEP empty braces:
\printAffiliationsAndNotice{}  % no special notice (required even if empty)
% Or, if applicable, use the standard equal contribution text:
% \printAffiliationsAndNotice{\icmlEqualContribution}

\begin{abstract}
Biological learning proceeds from easy to difficult tasks, gradually reinforcing perception and robustness. Inspired by this principle, we address Context‑Entangled Content Segmentation (CECS), a challenging setting where objects share intrinsic visual patterns with their surroundings, as in camouflaged object detection. Conventional segmentation networks predominantly rely on architectural enhancements but often ignore the learning dynamics that govern robustness under entangled data distributions.
We introduce CurriSeg, a dual‑phase learning framework that unifies curriculum and anti‑curriculum principles to improve representation reliability. In the Curriculum Selection phase, CurriSeg dynamically selects training data based on the temporal statistics of sample losses, distinguishing hard‑but‑informative samples from noisy or ambiguous ones, thus enabling stable capability enhancement. In the Anti‑Curriculum Promotion phase, we design Spectral‑Blindness Fine‑Tuning, which suppresses high‑frequency components to enforce dependence on low‑frequency structural and contextual cues and thus 
% . Combined with Sharpness‑Aware Minimization, this promotes convergence to flatter minima and 
strengthens generalization.
Extensive experiments demonstrate that CurriSeg achieves consistent improvements across diverse CECS benchmarks without adding parameters or increasing total training time, offering a principled view of how progression and challenge interplay to foster robust and context‑aware segmentation.
The code is available at \url{https://github.com/ChunmingHe/CurriSeg}.

\end{abstract}
\setlength{\abovedisplayskip}{2pt}
\setlength{\belowdisplayskip}{2pt}
\section{Introduction}
Context-Entangled Content Segmentation (CECS) aims to segment those extremely concealed objects that share intrinsic visual patterns with their surroundings~\cite{fan2020camouflaged,zhao2024spider,he2023strategic,lu2024mace,xiao2024survey,he2026ride}, with applications in medical analysis~\cite{tajbakhsh2015automated,hu2026evaluating,he2025diffusion}, autonomous perception~\cite{mei2020don,pmlr-v267-hu25e,tcsvt1,zheng2025towards}, etc.

\begin{figure}[t]
\setlength{\abovecaptionskip}{0cm}
	\centering
% \begin{minipage}{1\textwidth}
		\begin{subfigure}{0.09\textwidth}
		\centering
\includegraphics[width=\textwidth]{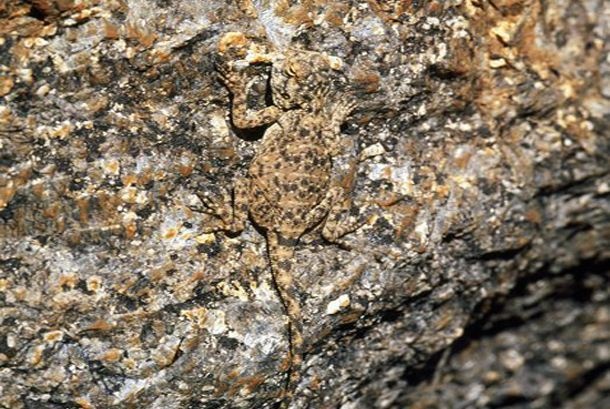}\vspace{-2pt}
	\end{subfigure}
	\begin{subfigure}{0.09\textwidth}  
		\centering 
\includegraphics[width=\textwidth]{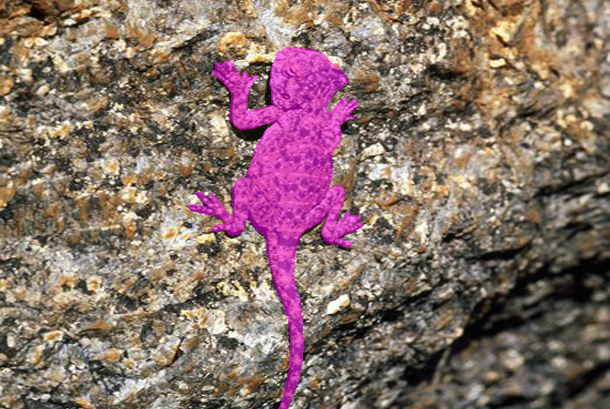}\vspace{-2pt}
	\end{subfigure}
	\begin{subfigure}{0.09\textwidth}  
		\centering 
\includegraphics[width=\textwidth]{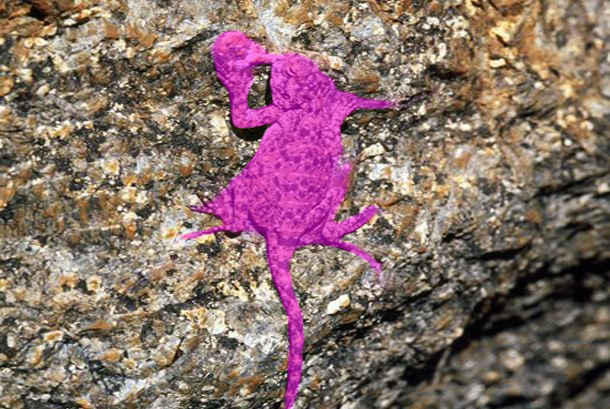}\vspace{-2pt}
	\end{subfigure}
    	\begin{subfigure}{0.09\textwidth} 
		\centering 
\includegraphics[width=\textwidth]{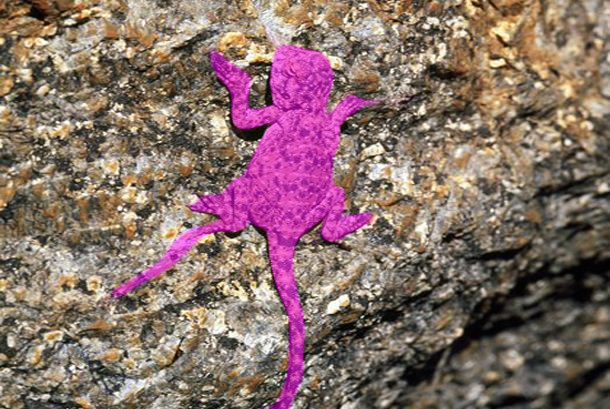}\vspace{-2pt}
	\end{subfigure}
	\begin{subfigure}{0.09\textwidth}  
		\centering 
\includegraphics[width=\textwidth]{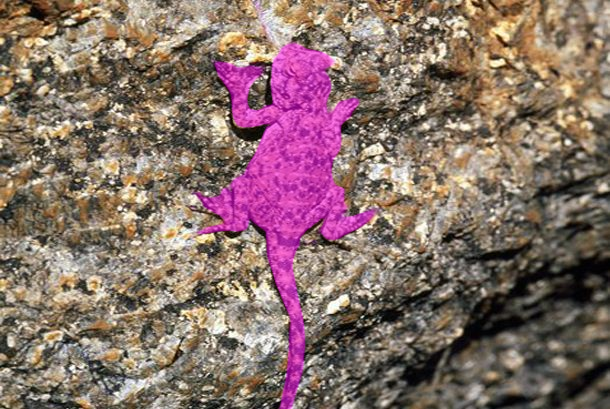}\vspace{-2pt}
	\end{subfigure} \\ \vspace{1mm}
\begin{subfigure}{0.09\textwidth}
\setlength{\abovecaptionskip}{0cm}
		\centering
\includegraphics[width=\textwidth]{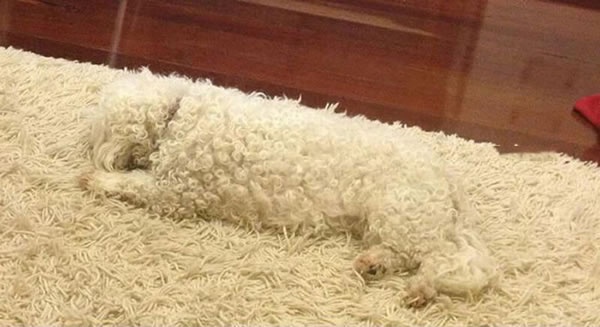} 
		\caption*{Origin}
	\end{subfigure}
	% \hfill
	\begin{subfigure}{0.09\textwidth} 
\setlength{\abovecaptionskip}{0cm}
		\centering 
\includegraphics[width=\textwidth]{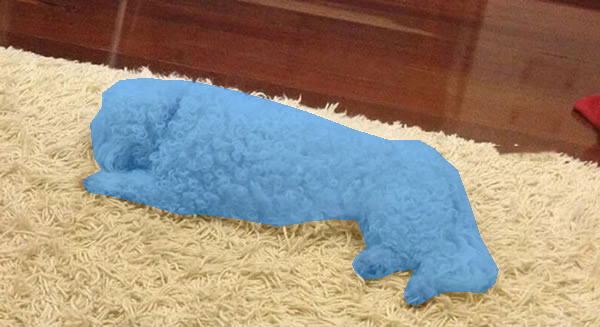} 
		\caption*{ GT }
	\end{subfigure}
	% \hfill
	\begin{subfigure}{0.09\textwidth}  
\setlength{\abovecaptionskip}{0cm}
		\centering 
\includegraphics[width=\textwidth]{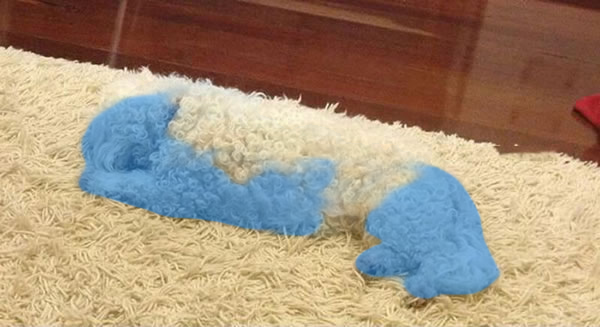} 
		\caption*{FEDER}
	\end{subfigure}
    % \hfill
    	\begin{subfigure}{0.09\textwidth} 
\setlength{\abovecaptionskip}{0cm}
		\centering 
\includegraphics[width=\textwidth]{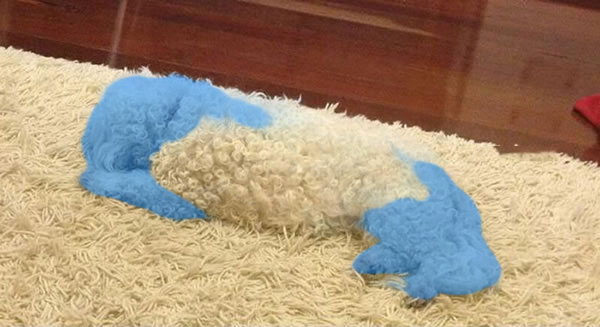} 
		\caption*{+ vanilla CL}
	\end{subfigure}
    % \hfill
	\begin{subfigure}{0.09\textwidth}  
\setlength{\abovecaptionskip}{0cm}
		\centering 
\includegraphics[width=\textwidth]{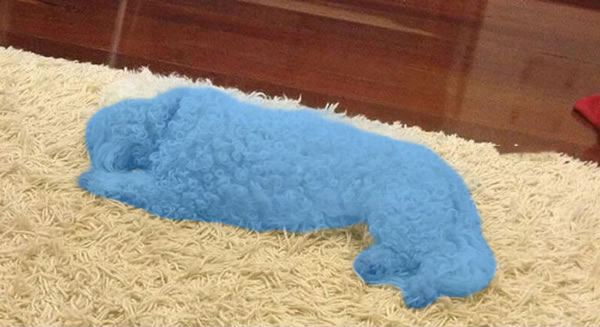} 
		\caption*{+ CurriSeg}
	\end{subfigure} \\
\begin{subfigure}{\linewidth}
		\centering
\includegraphics[width=\textwidth]{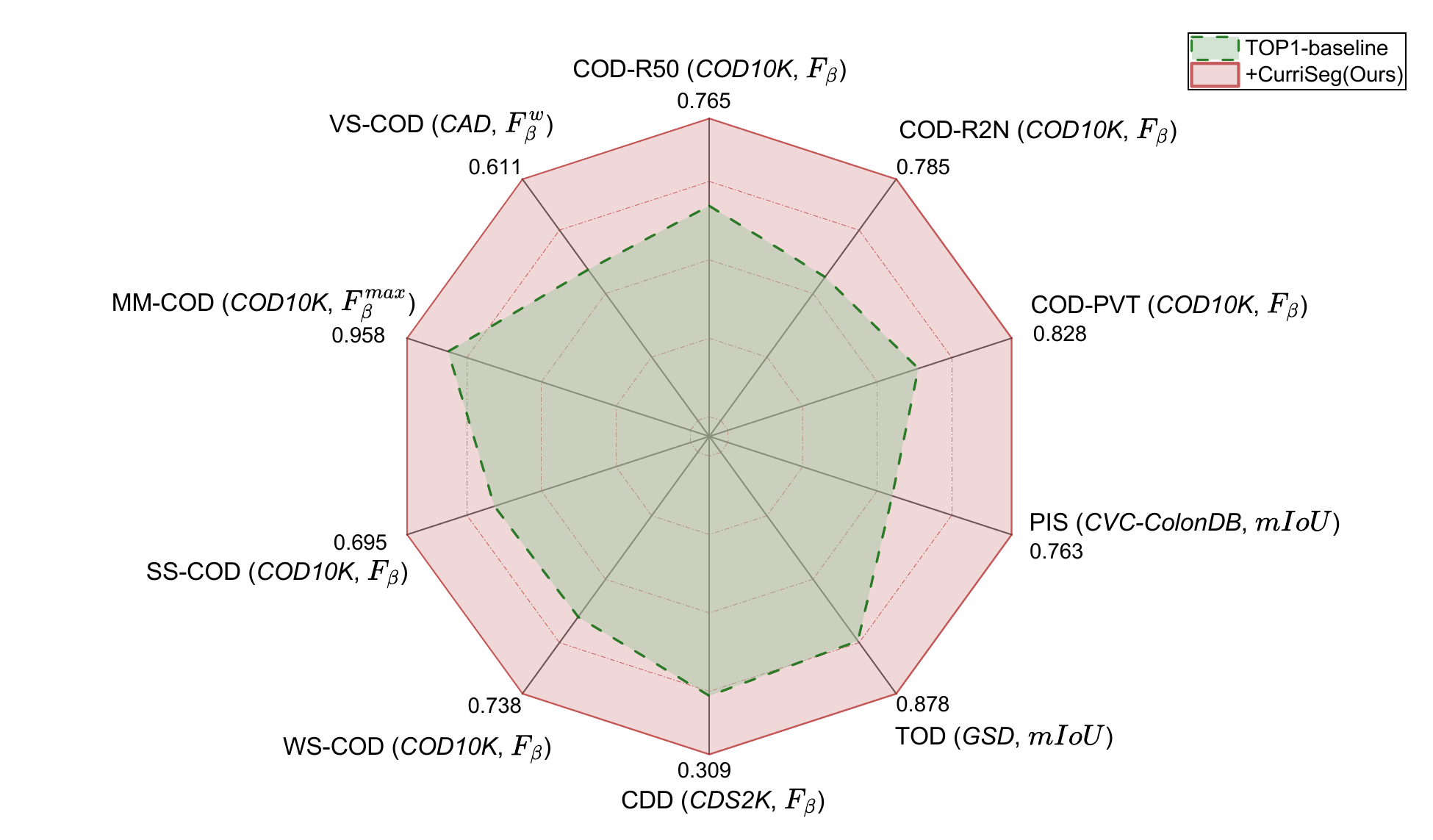} 
		% \caption*{Origin}
	\end{subfigure}
	\caption{Performance on CECS data. \textbf{Top}: CL is short for curriculum learning~\cite{bengio2009curriculum}. Concealed object masks are highlighted in {\color{purple}purple} and {\color{blue}blue}, overlaid on the original data for visual clarity. \textbf{Bottom}: ``Task (Dataset, Metric)''. The radar chart shows that training the baseline under our CurriSeg brings performance gains than the standard training manner, where TOP-1 is the composite baselines with the top metric scores per task. 
    }
	\label{fig:Intro}
    % \end{minipage}
	\vspace{-5mm}
\end{figure}

% Unlike generic segmentation, CECS suffers from extreme feature ambiguity. Existing methods predominantly focus on architectural engineering, such as feature aggregation~\cite{pang2022zoom,fan2020camouflaged} or complex decoders~\cite{he2025run,he2025reversible}. However, they overlook a fundamental bottleneck: the learning dynamics of the model when facing such an entangled data distribution.

Unlike generic segmentation, CECS suffers from extreme feature ambiguity. Existing methods mainly focus on architectural engineering, such as feature aggregation~\cite{pang2022zoom,fan2020camouflaged} complex decoders~\cite{he2025run,he2025reversible}, or prior injection~\cite{xiao2026beyond,xiao2026qualiteacher}, yet overlook a fundamental bottleneck: the learning dynamics when facing entangled data distributions.

% A natural strategy to handle such complexity is Curriculum Learning (CL)~\cite{bengio2009curriculum,soviany2022curriculum}, learning from easy to hard, mimicking biological cognition. However, we argue that standard CL is detrimental in CECS scenarios. In camouflaged scenes, ``easy'' samples often exhibit high spurious correlations (\textit{e.g.}, distinct background textures that unintentionally reveal the object). Training on these samples first without constraints drives the model into lazy regimes, where it relies on high-frequency texture bias rather than semantic structure, leading to poor generalization on hard samples (robustness failure). The performance shown in \cref{fig:Intro} also validates this. 

A natural strategy to handle such complexity is Curriculum Learning (CL)~\cite{bengio2009curriculum}, learning from easy to hard, mimicking biological cognition. However, we argue that standard CL is detrimental in CECS scenarios. In camouflaged scenes, ``easy'' samples often exhibit high spurious correlations (\textit{e.g.}, distinct background textures that unintentionally reveal the object). Training on these first drives the model into lazy regimes that rely on high-frequency texture bias rather than semantic structure, leading to poor generalization on hard samples, as validated in \cref{fig:Intro}.

To address this, we introduce CurriSeg, a novel framework that orchestrates training through two phases: Robust Curriculum Selection and Anti-Curriculum Promotion. Inspired by biological mastery, where predators first acquire fundamental skills then adapt to challenging environments, CurriSeg stabilizes optimization before deliberately ``blinding'' the model to encourage deeper feature extraction.

In the Curriculum Selection phase, we address the noise–ambiguity dilemma inherent in CECS. Instead of relying on static difficulty measures, we monitor temporal behavior of samples. By tracking mean and variance of sample losses, CurriSeg distinguishes hard-but-informative samples from noisy or ambiguous ones. We also incorporate pixel-level uncertainty estimation to prevent uncertain regions from dominating early gradients. Combined with a warm-up curriculum strategy, this yields a joint image- and pixel-level curriculum that is stable and noise-resistant.

After the model reaches a stable regime, we introduce Anti-Curriculum Promotion to enhance robustness on hard samples. We propose Spectral-Blindness Fine-Tuning (SBFT), which deliberately attenuates high-frequency input components. By temporarily suppressing texture cues, SBFT forces the model to rely on low-frequency structural patterns and contextual semantics, promoting extraction of intrinsic, task-relevant signals. This explicitly counteracts the tendency to exploit superficial texture shortcuts and encourages discovery of complementary, subtler cues, improving robustness where texture-based heuristics fail.

Together, these two phases establish a principled paradigm that stabilizes optimization then intentionally increases difficulty to overcome representational bottlenecks. CurriSeg combines curriculum and anti-curriculum learning within a unified framework, offering a new perspective on training for context-entangled visual tasks.

Our contributions are summarized as follows:

\noindent\textbf{(1)} We propose CurriSeg, the first CL-based CECS paradigm. By following a ``stabilize-then-perturb'' trajectory, it effectively breaks the learning bottleneck caused by feature ambiguity in concealed scenes.

\noindent\textbf{(2)} We design robust curriculum selection that uses temporal loss statistics and uncertainty-aware pixel masking to construct a stable, noise-resistant learning trajectory.

\noindent\textbf{(3)} We introduce an anti-curriculum promotion phase, SBFT, which attenuates high-frequency cues and encourages the model to exploit complementary, subtle signals, thereby improving robustness on challenging camouflaged cases.

\noindent\textbf{(4)} We validate CurriSeg on multiple CECS benchmarks and backbones, showing consistent performance gains. CurriSeg adds no extra parameters and maintains training cost comparable to standard training schedules.

\section{Related Works}
\noindent \textbf{Context-entangled content segmentation.}
Existing CECS methods mainly focus on structural designs: encoder-decoder backbones with multi-scale aggregation~\cite{fan2020camouflaged,fan2021concealed,Li2023FCDFusion}, attention-based refinement~\cite{pang2022zoom,pang2024zoomnext}, edge- or uncertainty-aware branches~\cite{He2023Camouflaged,he2025scaler}, transformer-style context modeling~\cite{sun2025frequency,zhang2025referring}, and frequency-aware modules~\cite{he2025nested,shen2025uncertainty}. They enhance feature representation at the network level, yet training typically follows a standard supervised routine with shuffled mini-batches. The roles of sample difficulty, label ambiguity, and texture shortcuts in shaping CECS learning dynamics remain underexplored. Our work instead organizes the training procedure via curriculum and anti-curriculum mechanisms, orthogonal to existing architectural advances.

\noindent \textbf{Curriculum learning}.
CL organizes training samples from easy to hard to improve generalization~\cite{bengio2009curriculum,soviany2022curriculum}. Variants include self-paced learning~\cite{kumar2010self,tullis2011effectiveness}, teacher–student curricula~\cite{matiisen2019teacher,saglietti2022analytical}, and task-specific schemes via loss-based ranking~\cite{pentina2015curriculum} or hardness-aware sampling~\cite{soviany2020curriculum}. However, CL assumes easy samples provide reliable supervision and difficulty correlates with learning utility. In CECS, these assumptions break down: ``easy'' samples may contain spurious textures promoting shortcuts, while hard samples can be both informative and ambiguous. Moreover, most CL methods overlook the pixel-level nature of segmentation with spatially localized uncertainty. CurriSeg departs from CL by (i) using temporal loss statistics to distinguish hard-but-informative samples from noisy cases, and (ii) adding an anti-curriculum phase that increases difficulty via spectral manipulation, mitigating texture bias.
% CL organizes training samples in a meaningful order, typically from easy to hard, to improve optimization and generalization~\cite{bengio2009curriculum,soviany2022curriculum}. Variants include self-paced learning~\cite{kumar2010self,tullis2011effectiveness}, teacher–student curricula~\cite{matiisen2019teacher,saglietti2022analytical}, and task-specific schemes in vision, often implemented via loss-based ranking~\cite{pentina2015curriculum} or hardness-aware sampling~\cite{soviany2020curriculum}. However, CL usually assumes that easy samples provide reliable supervision and that difficulty correlates with learning utility. In CECS, these assumptions break down: visually ``easy'' samples may contain spurious textures that promote shortcut solutions, while hard samples can be both informative and ambiguous. Moreover, most CL methods overlook the pixel-level nature of segmentation, where uncertainty is spatially localized. Our CurriSeg departs from standard CL by (i) using temporal loss statistics to distinguish hard-but-informative samples from noisy or outlier cases, and (ii) augmenting the curriculum with an anti-curriculum phase that deliberately increases difficulty via spectral manipulation to mitigate texture bias in CECS.

\begin{figure*}[h]
\setlength{\abovecaptionskip}{0cm}
	\centering
	\includegraphics[width=\linewidth]{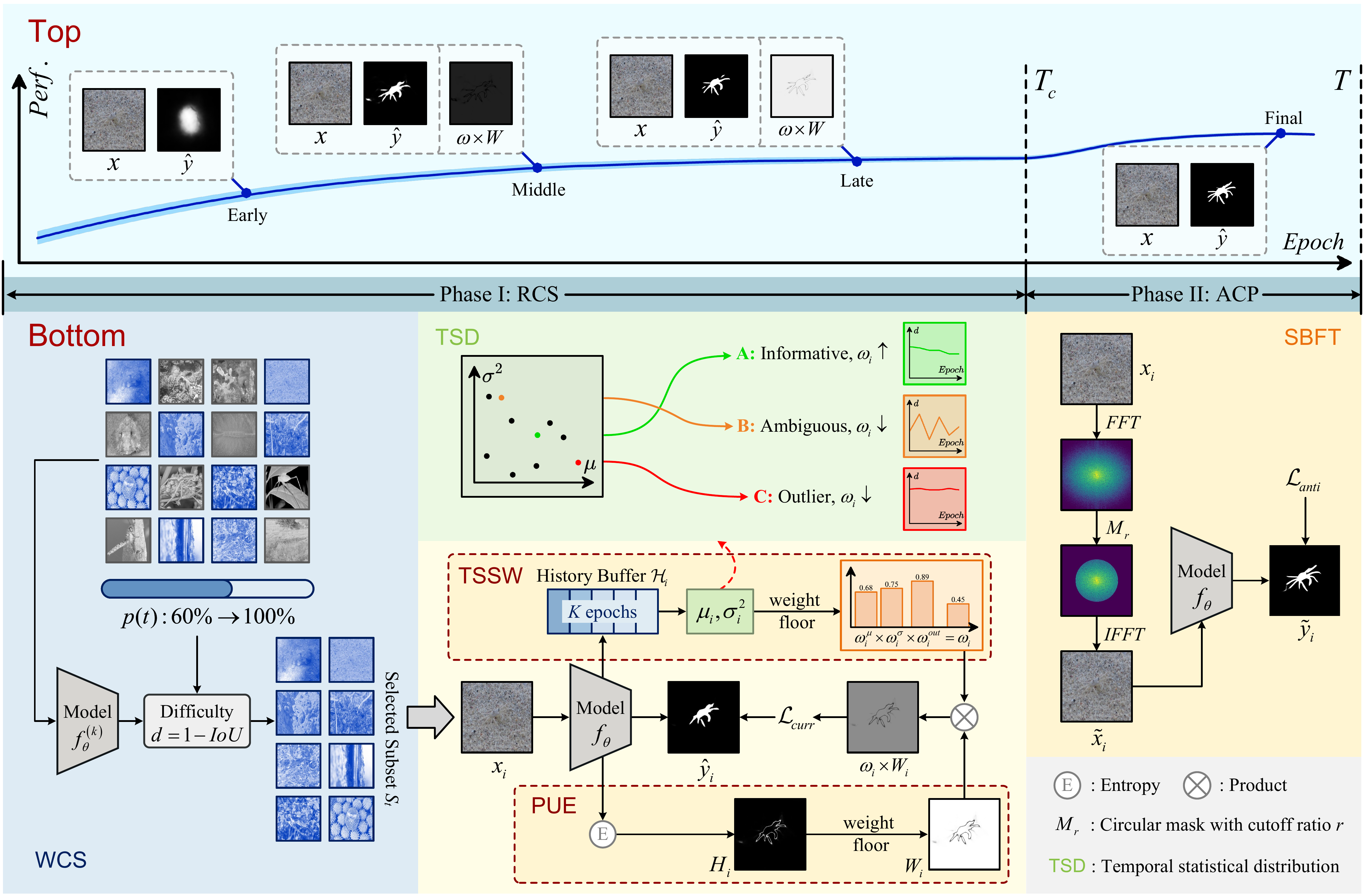}
    % \vspace{-4mm}
	\caption{Framework of CurriSeg. In (1), performance improves steadily; brighter regions in $\omega \times {W}$ indicate more trustworthy pixels.
    In (2), Phase I applies Robust Curriculum Selection with sample/pixel-level weighting, while Phase II includes Anti-Curriculum Promotion via spectral filtering. TSD describes how to distinguish hard-but-informative samples from ambiguous or outlier ones.
    }
	\label{fig:Framework}
	\vspace{-3mm}
\end{figure*}

\section{Methodology}
CurriSeg is a dual-phase learning framework designed to address the feature ambiguity inherent in CECS. As illustrated in \cref{fig:Framework,alg:curriseg}, the framework follows a ``stabilize-then-perturb'' trajectory comprising two sequential phases: (1) Robust Curriculum Selection (RCS), which constructs a stable, noise-resistant learning schedule through temporal statistics monitoring and uncertainty-aware weighting, and (2) Anti-Curriculum Promotion (ACP), which deliberately attenuates high-frequency cues to encourage the model to exploit complementary, subtle discriminative cues.

\subsection{Robust Curriculum Selection}
This phase targets the noise–ambiguity dilemma inherent in CECS by monitoring temporal behavior of samples and incorporating pixel-level uncertainty, yielding a joint image- and pixel-level curriculum for stable optimization.
% This phase targets the noise–ambiguity dilemma inherent in CECS. Unlike standard curriculum learning that relies on static difficulty measures, we monitor the temporal behavior of samples during training and incorporate segmentation-specific factors via pixel-level uncertainty. This yields a joint image-level and pixel-level curriculum that enables stable optimization under entangled data distributions.

\noindent\textbf{Warm-up curriculum strategy}.  
% Following~\cite{bengio2009curriculum}, we instantiate a warm-up curriculum strategy (WCS) that progressively expands the training set as the model becomes more competent. We adopt a model-centric notion of difficulty based on segmentation results, which adapts to the current state of the learner.
Following~\cite{bengio2009curriculum}, we design a warm-up curriculum strategy (WCS) that progressively expands the training set using a model-centric difficulty measure based on segmentation performance. Concretely, we maintain a historical checkpoint $f_{\theta^{(k)}}$ saved every $K$ epochs (we set $K=10$, so $k \in \{0,10,20,\cdots\}$). During the subsequent $K$ epochs, this checkpoint serves as an evaluator of sample difficulty. For each training sample $i$, we compute its difficulty score $d_i$, formulated as:
% as:
\begin{equation}
    d_i = 1-IoU(f_{\theta^{(k)}}(x_i),y_i),
\end{equation}
where $f_{\theta^{(k)}}(x_i)$ is the predicted mask of the concealed input $x_i$ from checkpoint model, $y_i$ is the ground-truth, and $IoU(\bigcdot)$ is the Intersection-over-Union. Samples with higher $d_i$ are considered more difficult for the current model. 

We then implement a warm-up curriculum schedule by gradually expanding the training set. Let $p(t)$ denote the selection percentile at epoch $t$, and let $\text{Per}(\{d_n\}_{n=1}^N,p(t))$ be the corresponding percentile threshold over all $N$ training samples. The active training subset at epoch $t$ is
\begin{equation}\label{eq:wcs}
\begin{aligned}
    &\mathcal{S}_t = \{i \ | \ d_i \leq \text{Per}(\{d_n\}_{n=1}^N,p(t))\}, \\
    &p(t)  = p_{min}+(1-p_{min})\cdot\frac{t-1}{T_c-1}, 
    % \quad 1\leq t \leq T_c,
\end{aligned}
\end{equation}
where $1\leq t \leq T_c$ and $p_{min}=0.6$.
In practice, training starts from a subset of relatively easy samples (small $p(t)$) and progressively incorporates more difficult ones as $p(t)$ increases. This encourages the model to first consolidate its understanding of easier cases before being exposed to the full complexity and ambiguity dataset.

\noindent\textbf{Temporal statistics-based sample weighting}.
Within the selected curriculum subset, not all samples contribute equally to effective learning. Due to object concealment, some samples are intrinsically ambiguous or affected by annotation noise; treating them on par with well-defined samples may destabilize optimization. To mitigate this, we exploit temporal statistics of sample-wise performance, termed TSSW, to down-weight problematic samples.
% Within the selected subset, some samples are innately ambiguous or annotation-noisy; treating them equally with well-defined samples may destabilize optimization. We exploit temporal statistics (TSSW) to down-weight such problematic samples.

% s. Also, owing to the concealment of the objects, certain samples may be inherently ambiguous or contain annotation errors. Treating such samples equally with well-defined ones can destabilize training. We propose to leverage temporal statistics of sample-wise performance to identify and down-weight problematic samples.

For each sample $i$, we maintain a circular buffer storing its difficulty scores over the past $K$ epochs: $\mathcal{H}_i = \{d_i^{(1)}, \cdots, d_i^{(K)}\}$, computing two temporal statistics:
\begin{equation}
    \mu_i= \frac{1}{K}\sum_{k=1}^K d_i^{(k)}, \ \ \ \sigma_i^2= \frac{1}{K}\sum_{k=1}^K (d_i^{(k)}-\mu_i)^2. 
\end{equation}
These statistics enable CurriSeg to distinguish hard-but-informative samples from noisy or ambiguous ones:
\begin{itemize}
\vspace{-3mm}
    \item \textbf{High variance} indicates that the model's predictions fluctuate significantly over time, suggesting the sample lies near the decision boundary or contains inherent ambiguity that destabilizes learning.
    \vspace{-3mm}
    \item \textbf{High mean error with low variance} indicates consistent failure, signaling a potential outlier or annotation error that the model cannot resolve.\vspace{-3mm}
\end{itemize}
By first applying min-max normalization to obtain $\tilde{\mu}_i$ and $\tilde{\sigma}_i^2$, we convert the temporal statistics into weights: 
\begin{equation}
\begin{aligned}
    &\omega_i^\mu=1-\tilde{\mu}_i, \ \omega_i^\sigma = \text{exp}\left(-\frac{(\tilde{\sigma}_i^2-\sigma^*)^2}{2\gamma^2} \right),    
    % \\ &\omega_i^{out}=1-\tilde{\mu}_i\cdot(1-\tilde{\sigma}_i^2),
\end{aligned}
\end{equation}
where $\sigma^*=0.5$ is the optimal variance level. $\gamma=0.2$ controls tolerance to variance deviation. The final weight is:
\begin{equation}
    \omega_i = W_{min}^s+(1-W_{min}^s)\cdot \omega_i^\mu \cdot \omega_i^\sigma \cdot (1-\tilde{\mu}_i\cdot(1-\tilde{\sigma}_i^2)),
\end{equation}
where $W_{min}^s=0.1$ provides a weight lower bound and $\cdot$ is product. We denote the last term as $\omega_i^{out}$, \textit{i.e.}, $\omega_i^{out}=1-\tilde{\mu}_i\cdot(1-\tilde{\sigma}_i^2)$. $\omega_i^{out}$ penalizes the ``outlier'' pattern of high mean error combined with low variance. Hence, CurriSeg emphasizes hard-but-informative samples while suppressing those that are unstable or likely mislabeled.

\noindent\textbf{Pixel-level uncertainty estimation}.
% Segmentation exhibits substantial within-sample variation in difficulty: pixels near boundaries or in low-contrast regions are inherently harder than those in homogeneous areas. If all pixels are treated equally, gradients from highly uncertain pixels (\textit{e.g.}, predictions near 0.5) can dominate and destabilize optimization. To address this, we introduce an entropy-based pixel weighting mechanism, named PUE, that attenuates the contribution of uncertain pixels while preserving informative gradients.
Segmentation exhibits within-sample variation: boundary and low-contrast pixels are harder than homogeneous ones. If treated equally, gradients from uncertain pixels can dominate optimization. We introduce entropy-based pixel weighting (PUE) to attenuate uncertain pixels while preserving informative gradients.

In contrast to prior work that typically uses uncertainty to filter pseudo labels~\cite{he2023weaklysupervised,he2025segment,he2025scaler}, we apply uncertainty directly on the predicted mask. The goal is not to hide failure regions from the model, but to prevent them from disproportionately influencing learning.

For a prediction logit $\hat{y}_{h,w}$ at spatial location $(h, w)$, we compute its predicted probability $p_{h,w}$ with Sigmoid $\sigma$: $p_{h,w}=\sigma(\hat{y}_{h,w})$. The normalized prediction entropy is:
\begin{equation}
    H_{h,w}=-p_{h,w}\log_2 p_{h,w} - (1-p_{h,w})\log_2 (1-p_{h,w}).
\end{equation}
$H_{h,w} \in [0, 1]$ reaches its maximum when $p_{h,w} = 0.5$ (maximum uncertainty) and minimum when $p_{h,w} \in \{0, 1\}$ (complete certainty). Instead of hard masking, we define a soft weighting matrix ${W}_i$ whose entry at $(h,w)$ and epoch $t$ is:
\begin{equation}
    W_{h,w}(t) = W_{min} + (1-W_{min})\cdot (1-\beta(t)\cdot H_{h,w}),
\end{equation}
where $W_{min}=0.1$ sets a non-zero weight floor, and $\beta(t)=(1-t/T_c)$ is a curriculum-aware coefficient that decays over the curriculum phase of length $T_c$. Early in training, high-entropy pixels receive substantially reduced weights, limiting gradient noise from ambiguous regions. As training progresses and $\beta(t)$ decreases, full supervision is gradually restored, while the weight floor ensures continuous, albeit reduced, gradient flow from all pixels.

\begin{table*}[tbp!]
\begin{minipage}{1\textwidth}
\setlength{\abovecaptionskip}{0cm} 
		\centering
            \caption{Results on camouflaged object detection. The suffix ``+'' indicates that the network is trained under our CurriSeg. $\Delta$: increase.
            } \label{table:CODQuanti}
            \vspace{1mm}
		\resizebox{\columnwidth}{!}{
			\setlength{\tabcolsep}{1.4mm}
			\begin{tabular}{l|c|cccc|cccc|cccc|cccc|c} 
				\toprule
				\multicolumn{1}{c|}{}                                        & \multicolumn{1}{c|}{}                           & \multicolumn{4}{c|}{\textit{CHAMELEON} }                                                                                                                                         & \multicolumn{4}{c|}{\textit{CAMO} }                                                                                                                                             & \multicolumn{4}{c|}{\textit{COD10K} }                                                                                                                                          & \multicolumn{4}{c|}{\textit{NC4K} }  & \textbf{COD}                                                                                                                      \\ \cline{3-19} 
				\multicolumn{1}{l|}{\multirow{-2}{*}{Methods}} & \multicolumn{1}{c|}{\multirow{-2}{*}{Backbones}} & {\cellcolor{gray!40}$M$~$\downarrow$}                                  & {\cellcolor{gray!40}$F_\beta$~$\uparrow$}                               & {\cellcolor{gray!40}$E_\phi$~$\uparrow$}                               & \multicolumn{1}{c|}{\cellcolor{gray!40}$S_\alpha$~$\uparrow$}                                   & {\cellcolor{gray!40}$M$~$\downarrow$}                                  & {\cellcolor{gray!40}$F_\beta$~$\uparrow$}                               & {\cellcolor{gray!40}$E_\phi$~$\uparrow$}                               & \multicolumn{1}{c|}{\cellcolor{gray!40}$S_\alpha$~$\uparrow$}                                   & {\cellcolor{gray!40}$M$~$\downarrow$}                                  & {\cellcolor{gray!40}$F_\beta$~$\uparrow$}                               & {\cellcolor{gray!40}$E_\phi$~$\uparrow$}                               & \multicolumn{1}{c|}{\cellcolor{gray!40}$S_\alpha$~$\uparrow$}                                   & {\cellcolor{gray!40}$M$~$\downarrow$}                                  & {\cellcolor{gray!40}$F_\beta$~$\uparrow$}                               & {\cellcolor{gray!40}$E_\phi$~$\uparrow$}                               & \multicolumn{1}{c|}{\cellcolor{gray!40}$S_\alpha$~$\uparrow$} & $\Delta$ (\%)                             \\ \midrule 
                \multicolumn{1}{l|}{SINet~\cite{fan2020camouflaged}}& \multicolumn{1}{c|}{ResNet50}                   & 0.034                                 & 0.823                                 & 0.936                                 & \multicolumn{1}{c|}{0.872}                                 & 0.092                                 & 0.712                                 & 0.804                                 & \multicolumn{1}{c|}{0.745}                                 & 0.043                                 & 0.667                                 & 0.864                                 & \multicolumn{1}{c|}{0.776}                                 & 0.058                                 & 0.768                                 & 0.871                                 & 0.808   & ---                              \\
				\multicolumn{1}{l|}{MGL-R~\cite{zhai2021mutual}}                     & \multicolumn{1}{c|}{ResNet50}                   & 0.031                                 & 0.825                                 & 0.917                                 & \multicolumn{1}{c|}{0.891}                                 & 0.088                                 & 0.738                                 & 0.812                                 & \multicolumn{1}{c|}{0.775}                                 & 0.035                                 & 0.680                                 & 0.851                                 & \multicolumn{1}{c|}{0.814}                                 & 0.053                                 & 0.778                                 & 0.867                                 & 0.833   & ---                               \\
				\multicolumn{1}{l|}{PreyNet~\cite{zhang2022preynet}}                     & \multicolumn{1}{c|}{ResNet50}                   & {{0.027}}  & 0.844 & {{0.948}}  & \multicolumn{1}{c|}{0.895}   & 0.077   & 0.763   & 0.854   & \multicolumn{1}{c|}{0.790}  & 0.034   & 0.715  &{{0.894}}  & \multicolumn{1}{c|}{0.813}                                 & 0.047   & 0.798                                 & 0.887                                 & 0.838      & ---                            \\
                \multicolumn{1}{l|}{FEDER~\cite{He2023Camouflaged}} & \multicolumn{1}{c|}{ResNet50}  & {{0.028}} & {{0.850}} & 0.944 & \multicolumn{1}{c|}{0.892} & {{0.070}} & {0.775} & 0.870 & \multicolumn{1}{c|}{0.802} & 0.032 & 0.715 & 0.892 & \multicolumn{1}{c|}{0.810} & {{0.046}} & {{0.808}} & {{0.900}} & {{0.842}} & --- \\
                \rowcolor{c2!20} FEDER+ (Ours) & ResNet50 & 0.026 & 0.858 & 0.952 & 0.898 & 0.068 & 0.790 & 0.881 & 0.807 & 0.030 & 0.736 & 0.910 & 0.818 & 0.043 & 0.825 & 0.912 & 0.850 & 2.46 $\uparrow$   \\
                \multicolumn{1}{l|}{FSEL~\cite{sun2025frequency}} & \multicolumn{1}{c|}{ResNet50} & 0.029 & 0.847 & 0.941 & {{0.893}}  & {{{0.069}}} & {{0.779}} & {0.881} & {{{0.816}}} & 0.032 & 0.722 & 0.891 & {{0.822}} & 0.045 & 0.807 & 0.901 & 0.847  & ---   \\ 
                \rowcolor{c2!20} FSEL+ (Ours) & ResNet50 & 0.028 & 0.856 & 0.949 & 0.898 & \textbf{0.067} & 0.792 & \textbf{0.889} & \textbf{0.819} & 0.030 & 0.742 & 0.909 & 0.831 & 0.042 & 0.823 & 0.919 & 0.855 & 2.22 $\uparrow$  \\
                \multicolumn{1}{l|}{RUN~\cite{he2025run}}  & \multicolumn{1}{c|}{ResNet50} & {{0.027}} & {{0.855}} & {{0.952}} & {{0.895}} & 0.070 & {{0.781}} & 0.868 & 0.806 & {0.030} & {{0.747}} & {{0.903}} & {{0.827}} & {0.042} & {{0.824}} & {{0.908}} & {{0.851}} & ---     \\
                \rowcolor{c2!20} RUN+ (Ours) & ResNet50 & \textbf{0.025} & \textbf{0.863} & \textbf{0.960} & \textbf{0.900} & 0.068 & \textbf{0.801} & 0.879 & 0.813 & \textbf{0.029} & \textbf{0.765} & \textbf{0.917} & \textbf{0.836} & \textbf{0.040} & \textbf{0.835} & \textbf{0.921} & \textbf{0.858} & 2.13 $\uparrow$  \\
                \midrule
				\multicolumn{1}{l|}{BSA-Net~\cite{zhu2022can}}            & \multicolumn{1}{c|}{Res2Net50}                  & 0.027                                 & 0.851                                 & 0.946                                 & \multicolumn{1}{c|}{0.895}                                 & 0.079                                 & 0.768                                 & 0.851                                 & \multicolumn{1}{c|}{0.796}                                 & 0.034                                 & 0.723                                 & 0.891                                 & \multicolumn{1}{c|}{0.818}                                 & 0.048                                 & 0.805                                 & 0.897                                 & 0.841            & ---                      \\
                % \rowcolor{c2!20} BSA-Net+ (Ours) & Res2Net50 &   \\
               \multicolumn{1}{l|}{RUN~\cite{he2025run}}  & \multicolumn{1}{c|}{Res2Net50} & {0.024} & {0.879} & {0.956} & {0.907} & {{0.066}} & {0.815} & {0.905} & {{0.843}}  & {0.028} & {0.764} & {{0.914}} & {{0.849}} & {0.041} & {0.830} & {0.917} &0.859     & ---            \\
               \rowcolor{c2!20} RUN+ (Ours) & Res2Net50 & \textbf{0.023} & \textbf{0.891} & \textbf{0.963} & \textbf{0.911} & \textbf{0.065} & \textbf{0.820} & \textbf{0.912} & \textbf{0.845} & \textbf{0.026} & \textbf{0.785} & \textbf{0.933} & \textbf{0.857} & \textbf{0.038} & \textbf{0.852} & \textbf{0.932} & \textbf{0.870} & 2.23 $\uparrow$  \\
                \midrule
                % \multicolumn{1}{l|}{CamoFocus~\cite{khan2024camofocus} }               & \multicolumn{1}{c|}{PVT V2} & 0.023 & 0.869 & 0.953 & 0.906 & {0.044} & {0.861} & 0.924 &0.870 & 0.022 & {0.818} & 0.931 & 0.868 & 0.031 & 0.862 &0.932 &0.886 \\
                CamoDiff~\cite{sun2025conditional} & PVT V2 & 0.022 & 0.868 & 0.952 & 0.908 & {0.042} & 0.853 & 0.936 & 0.878 & {0.019} & 0.815 & 0.943 & 0.883 & {0.028} & 0.858 & 0.942 & 0.895  & ---  \\
                % \rowcolor{c2!20} CamoDiff+ (Ours) & PVT V2 &   \\
                \multicolumn{1}{l|}{RUN~\cite{he2025run}} &  \multicolumn{1}{c|}{PVT V2} & {{0.021}} & {0.877} & {{0.958}} & {{0.916}} & 0.045 &  {0.861} &  {0.934} & {0.877} & {0.021} &  0.810 & {0.941} &  {0.878} & {0.030} & {0.868} & {0.940} & {0.892}  & ---  \\ 
                \rowcolor{c2!20} RUN+ (Ours) & PVT V2 & \textbf{0.019} & \textbf{0.893} & \textbf{0.971} & \textbf{0.922} & \textbf{0.042} & \textbf{0.879} & \textbf{0.952} & \textbf{0.885} & \textbf{0.018} & \textbf{0.828} & \textbf{0.957} & \textbf{0.886} & \textbf{0.026} & \textbf{0.889} & \textbf{0.958} & \textbf{0.903} & 3.94 $\uparrow$  \\
                \bottomrule            
		\end{tabular}}  
\end{minipage}
\vspace{-3mm}
	\end{table*} 
\begin{table}[ht]
\setlength{\abovecaptionskip}{0cm} 
\centering
\caption{Training overhead analysis (batch size: 2).
        } \label{table:overhead}
	\resizebox{1\columnwidth}{!}{
		\setlength{\tabcolsep}{2.1mm}
\begin{tabular}{l|cc|cc|cc}\toprule 
Metrics    & FEDER & \cellcolor{c2!20} FEDER+ & FSEL & \cellcolor{c2!20} FSEL+  & RUN & \cellcolor{c2!20} RUN+   \\ \midrule
Training Time (h)  & 9.62   &\cellcolor{c2!20}  6.84    & 11.54 &\cellcolor{c2!20}  5.96    & 12.64 &\cellcolor{c2!20}  8.32    \\
GPU Mem (G)     & 1.53   &\cellcolor{c2!20}  1.62    & 2.83  &\cellcolor{c2!20}  2.92    & 3.66  &\cellcolor{c2!20}  3.75    \\
Perf. Gain (\%) & ---   &\cellcolor{c2!20}  2.46 $\uparrow$ & ---  &\cellcolor{c2!20}  2.22 $\uparrow$ & ---  &\cellcolor{c2!20}  2.13 $\uparrow$ \\ \bottomrule
\end{tabular}}
		\vspace{-0.3cm}
	\end{table} 
\begin{table}[ht]
\setlength{\abovecaptionskip}{0cm} 
\centering
\caption{Results on polyp image segmentation.
        } \label{table:PISQuanti}
	\resizebox{1\columnwidth}{!}{
		\setlength{\tabcolsep}{0.5mm}
	\begin{tabular}{l|ccc|ccc|c}
		\toprule 
		\multirow{2}{*}{Methods} & \multicolumn{3}{c|}{\textit{CVC-ColonDB} }  & \multicolumn{3}{c|}{\textit{ETIS} } &  \textbf{PIS}
        \\ \cline{2-8} 
		& \multicolumn{1}{c}{\cellcolor{gray!40}mDice~$\uparrow$} & \multicolumn{1}{c}{\cellcolor{gray!40}mIoU~$\uparrow$} & \multicolumn{1}{c|}{\cellcolor{gray!40}$S_\alpha$~$\uparrow$} & \multicolumn{1}{c}{\cellcolor{gray!40}mDice~$\uparrow$} & \multicolumn{1}{c}{\cellcolor{gray!40}mIoU~$\uparrow$} & \multicolumn{1}{c|}{\cellcolor{gray!40}$S_\alpha$~$\uparrow$} & $\Delta$ (\%) \\ \midrule
        % CASCADE~\cite{rahman2023medical} & 0.809 & 0.731 & 0.867 & 0.781 & {{0.706}} & 0.853  \\
        PolypPVT~\cite{dong2023polyp}  & 0.808 &0.727 & 0.865 & {{0.787}} & {{0.706}} & {{0.871}} & --- \\
        CoInNet~\cite{jain2023coinnet} & 0.797 &0.729 &{{0.875}} & 0.759 & 0.690 & 0.859 & --- \\
        LSSNet~\cite{wang2024lssnet} & {{0.820}} & {{0.741}} & 0.867 & 0.779 & 0.701 & 0.867 & --- \\
        MEGANet~\cite{bui2024meganet} & 0.793 & 0.714 & 0.854 & 0.739 & 0.665 & 0.836 & --- \\
        \rowcolor{c2!20} MEGANet+ (Ours) & 0.815 & 0.736 & 0.866 & 0.755 & 0.682 & 0.848 & 2.24 $\uparrow$   \\
        RUN~\cite{he2025run} & {{0.822}} & {{0.742}} & {{0.880}} & {{0.788}} &{{0.709}} & {{0.878}} & ---  \\
        \rowcolor{c2!20} RUN+ (Ours) & \textbf{0.845} & \textbf{0.763} & \textbf{0.893} & \textbf{0.806} & \textbf{0.720} & \textbf{0.890} & 2.05 $\uparrow$   \\
	 \bottomrule                      
	\end{tabular}}\label{table:polyp}
		\vspace{-0.2cm}
	\end{table} 

\noindent\textbf{Unified curriculum loss}. Our overall objective follows standard CECS practice~\cite{fan2020camouflaged,He2023Camouflaged}, augmented with the proposed curriculum mechanisms. For a sample $i\in S_t$ at epoch $t$, the loss is defined as:
\begin{equation}\hspace{-3mm}
    \mathcal{L}_{curr}= \omega_i \cdot \left[\! L^w_{BCE}\!\left(\hat{y}_i,y_i,{W}_i\right)\!+\!L^w_{IoU}\!\left(\hat{y}_i,y_i,{W}_i\right)\right],
\end{equation}
where $\omega_i$ is the sample-level weight from temporal statistics. $L_{BCE}^w$ and $L_{IoU}^w$ are the weighted binary cross-entropy loss and weighted intersection-over-union loss with the combined pixel-level weight ${W}_i$. This joint image- and pixel-level curriculum yields a stable
% and noise-resistant 
learning schedule.
% tailored to concealed scenes.

\subsection{Anti-Curriculum Promotion}
Once the model has reached a stable representation regime, we introduce an Anti-Curriculum Promotion phase to enhance its performance on hard samples. This phase counteracts the tendency of the network to exploit superficial texture shortcuts, an inherent bias in dense prediction, and encourages the discovery of complementary, subtler cues.

% We propose Spectral-Blindness Fine-Tuning (SBFT), in which the high-frequency components of the input are deliberately attenuated. By suppressing fine-grained texture information, SBFT forces the model to rely more heavily on low-frequency structural patterns and contextual semantics, thereby promoting the extraction of more intrinsic, task-relevant discriminative signals. 

% Once the model reaches a stable regime, we introduce Anti-Curriculum Promotion to enhance performance on hard samples by counteracting texture shortcuts.

We propose Spectral-Blindness Fine-Tuning (SBFT), which attenuates high-frequency input components. By suppressing texture information, SBFT forces reliance on low-frequency structural patterns and contextual semantics, promoting extraction of intrinsic, task-relevant signals.

Given the concealed input $x_i$, we apply a low-pass filter in the frequency domain, formulated as:
\begin{equation}\label{eq:SBFT1}
    \tilde{x}_i=\mathcal{F}^{-1}(\mathcal{F}(x_i)\odot M_r),
\end{equation}
where $\mathcal{F}$ and $\mathcal{F}^{-1}$ denote the 2D Fourier transform and its inverse, and $M_r$ is a circular mask that preserves frequency components within a radius ratio $r$, defined as:
\begin{equation}\hspace{-3mm}
    M_r(u,v)\!=\!\mathbf{1}\!\left[ (u \!-\! c_u)^2 \!+\! (v \!-\! c_v)^2 \!\leq\!  \frac{r \!\cdot\!\min(H,W)}{2}  \right]\!,
\end{equation}
where $\mathbf{1}(\bigcdot)$ is a indicator function, $H,W$ are the image height and width, $(c_u, c_v)$ is the frequency domain center and $r = 0.95$ controls the cutoff ratio.
\begin{table}[ht]
	\setlength{\abovecaptionskip}{0cm} 
	\setlength{\belowcaptionskip}{-0.2cm}
	\centering
        \caption{Results on transparent object detection. 
        % Following the common practice, we employ ResNext101 as our backbone.
        } \label{table:TODQuanti}
	\resizebox{\columnwidth}{!}{
		\setlength{\tabcolsep}{0.5mm}
	\begin{tabular}{l|ccc|ccc|c}\toprule 

		\multicolumn{1}{l|}{}                          & \multicolumn{3}{c|}{\textit{GDD} } & \multicolumn{3}{c|}{\textit{GSD} } & \textbf{TOD}  \\ \cline{2-8} 
		\multicolumn{1}{l|}{\multirow{-2}{*}{Methods}} & \cellcolor{gray!40}mIoU~$\uparrow$&\cellcolor{gray!40}$F_\beta^{max}$~$\uparrow$&\cellcolor{gray!40} $M$~$\downarrow$& \cellcolor{gray!40}mIoU~$\uparrow$&\cellcolor{gray!40}$F_\beta^{max}$~$\uparrow$&\cellcolor{gray!40} $M$~$\downarrow$ & $\Delta$ (\%) \\ \midrule
		EBLNet~\cite{he2021enhanced} & 0.870                                 & 0.922                                 & 0.064                                  & 0.817                                 & 0.878                                 & 0.059 & --- \\
        IEBAF~\cite{han2023internal} & 0.887 & {{0.944}} & 0.056  &  0.861 & 0.926 & 0.049 & --- \\
        GhostingNet~\cite{yan2024ghostingnet} & {{0.893}}  & 0.943 & {{0.054}} & 0.838 & 0.904 & 0.055 & ---  \\
        RFENet~\cite{fan2023rfenet} & 0.886 & 0.938 & 0.057 & {{0.865}} & {{0.931}} & {{0.048}} & --- \\
        \rowcolor{c2!20} RFENet+ (Ours) & 0.893 & 0.952 & 0.052 & 0.875 & 0.939 & 0.045 & 3.22 $\uparrow$   \\
	RUN~\cite{he2025run}  & {{0.895}} & {{0.952}} & {{0.051}} & {{0.866}} & {{0.938}} & {{0.043}} & ---  \\ 
    \rowcolor{c2!20} RUN+ (Ours) & \textbf{0.902} & \textbf{0.966} & \textbf{0.047} & \textbf{0.878} & \textbf{0.945} & \textbf{0.039} & 3.59 $\uparrow$   \\
    \bottomrule  \end{tabular}}
\vspace{-2mm}
\end{table}By attenuating high-frequency texture details, SBFT creates an information bottleneck that reduces reliance on superficial patterns; the model is instead driven to exploit low-frequency boundaries, semantic context, and subtle intensity variations, which tend to be more robust across diverse concealed scenarios and thus enhance the ability of the model in handling hard cases.

Let $\tilde{y}$ denote the predicted mask for $\tilde{x}_i$. The loss function is:
\begin{equation}\hspace{-3mm}
    \mathcal{L}_{anti}= L^w_{BCE}\!\left(\tilde{y}_i,y_i\right)\!+\! L^w_{IoU}\!\left(\tilde{y}_i,y_i\right).
\end{equation}

\begin{figure*}[t]
\centering
\setlength{\abovecaptionskip}{0cm}
\includegraphics[width=\linewidth]{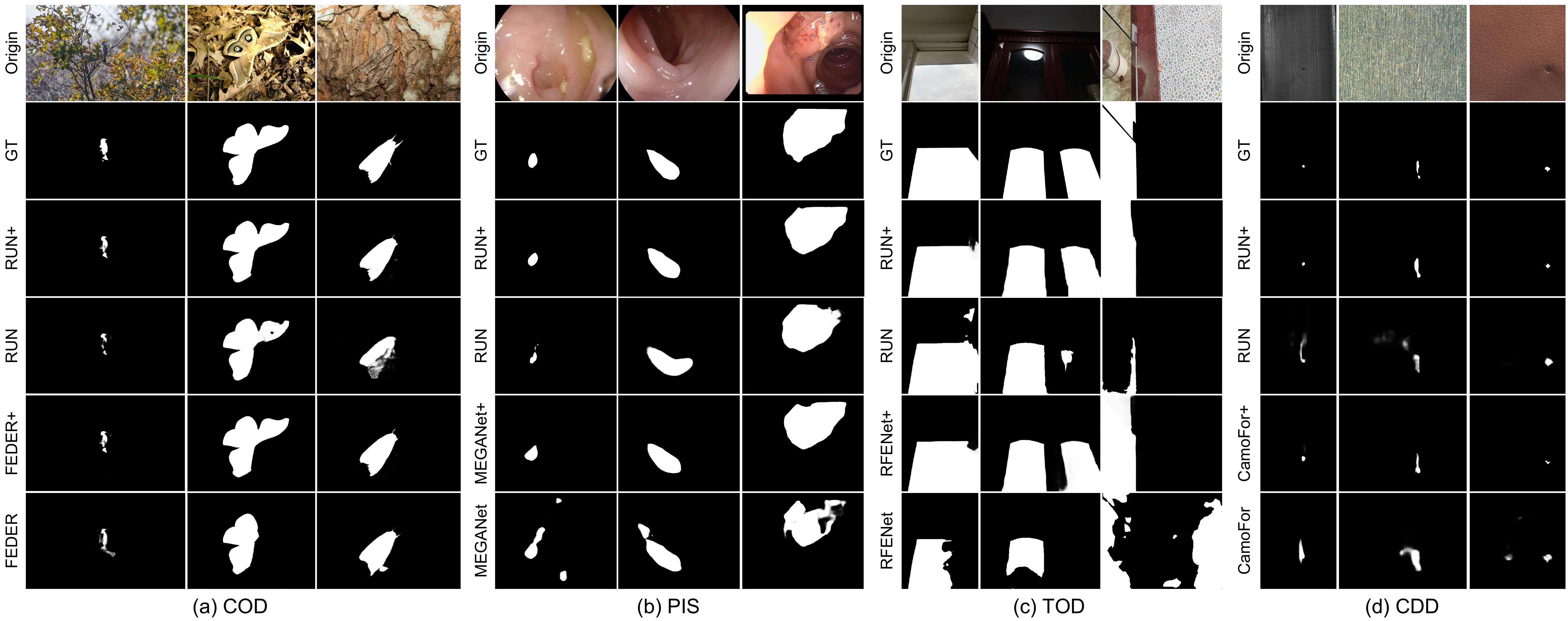}
\caption{Visual comparison on COD, PIS, TOD, and CDD tasks.}
\label{fig:visualization}
\vspace{-4mm}
\end{figure*}

% \vspace{-2mm}
\subsection{Overall Training Pipeline}
% \vspace{-1mm}
\cref{alg:curriseg} summarizes the CurriSeg training procedure:
% . The framework operates in two sequential phases:
\begin{itemize}
\vspace{-3mm}
    \item \textbf{Phase 1 (Epochs 1 to $T_c$):} Robust Curriculum Selection builds stable representations through progressive sample introduction, temporal statistics-based weighting, and pixel-level uncertainty estimation.
    \vspace{-3mm}
    \item \textbf{Phase 2 (Epochs $T_c$ to $T$):} Anti-Curriculum Promotion enhances robustness to hard cases through SBFT.\vspace{-3mm}
\end{itemize}
\vspace{-2mm}
CurriSeg does not modify the model architecture and introduces no extra parameters at inference. The overall training cost remains comparable to standard schedules, while the curriculum phase often accelerates effective convergence by focusing computation on more informative samples.

\begin{table*}[ht]
 \begin{minipage}[c]{0.47\textwidth}
	\centering
	\setlength{\abovecaptionskip}{0cm}
	\caption{Results on concealed defect detection.  
    % We employ PVT V2 as our backbone.
    } \label{table:CDDQuanti}
	\resizebox{1\columnwidth}{!}{
		\setlength{\tabcolsep}{1.4mm}
		\begin{tabular}{l|ccccc|c}
        \toprule 
Methods    & {\cellcolor{gray!40}$S_\alpha$~$\uparrow$} & {\cellcolor{gray!40}$M$~$\downarrow$} & {\cellcolor{gray!40}$E_\phi$~$\uparrow$} & {\cellcolor{gray!40}$F_\beta$~$\uparrow$} & {\cellcolor{gray!40}$F_\beta^{mean}$~$\uparrow$} &  $\Delta$ (\%)  \\ \midrule
SINet V2~\cite{fan2021concealed}   & 0.551 & 0.102  & 0.567   & 0.223  & 0.248 & ---  \\
HitNet~\cite{hu2022high}     & 0.563 & 0.118   & 0.564   & {{0.298}}  & 0.298 & ---  \\
OAFormer~\cite{yang2023oaformer}   & 0.541 & 0.121  & 0.535 & 0.216  & 0.239 & ---  \\
CamoFormer~\cite{yin2024camoformer} & {{0.589}} & {{0.100}} & {{0.588}}   & {{0.330}}  & {{0.329}} & ---   \\
\rowcolor{c2!20} CamoFormer+ (Ours) & \textbf{0.599} & 0.091 & 0.603 & \textbf{0.345} & \textbf{0.346} & 4.59 $\uparrow$   \\
RUN~\cite{he2025run} & {{0.590}} &  {{0.068}}  & {{0.595}}  & {{0.298}}  & {{0.299}} & ---  \\
    \rowcolor{c2!20} RUN+ (Ours) & {0.598} & \textbf{0.061} & \textbf{0.612} &  {0.309} &  {0.310} & 4.38 $\uparrow$  \\
\bottomrule
\end{tabular}}
	% \vspace{-0.3cm}
  \end{minipage}
% \end{table}
% \begin{table}[ht]
 \begin{minipage}[c]{0.508\textwidth}
	\centering
	\setlength{\abovecaptionskip}{0cm}
	\caption{Breakdown ablations of CurriSeg.  
    % We employ PVT V2 as our backbone.
    } \label{table:breakdown}
	\resizebox{1\columnwidth}{!}{
		\setlength{\tabcolsep}{3 mm}
		\begin{tabular}{cccccccc} \toprule
WCS                  & PUE                  & TSSW & SBFT & $M$~$\downarrow$ & $F_\beta$~$\uparrow$ & $E_\phi$~$\uparrow$ & $S_\alpha$~$\uparrow$    \\ \midrule
$\times$ & $\times$ & $\times$ & $\times$ & 0.032          & 0.715          & 0.892          & 0.810          \\
$\checkmark$  & $\times$ & $\times$ & $\times$  & 0.035          & 0.697          & 0.870          & 0.801          \\
$\checkmark$  & $\checkmark$ & $\times$ & $\times$  & 0.033          & 0.718          & 0.895          & 0.809          \\
$\checkmark$  & $\checkmark$ & $\checkmark$ & $\times$   & 0.031          & 0.729          & 0.904          & 0.815          \\
$\times$ & $\times$ & $\times$ & $\checkmark$ & 0.031          & 0.723          & 0.902          & 0.813          \\
\rowcolor{c2!20} $\checkmark$  & $\checkmark$ & $\checkmark$  & $\checkmark$ &  \textbf{0.030} & \textbf{0.736} & \textbf{0.910} & \textbf{0.818} \\ \bottomrule
\end{tabular}}
	% \vspace{-0.3cm}
  \end{minipage}\\
% \end{table*}
% \begin{table*}[ht]
% \centering
\begin{minipage}{1\textwidth}
\setlength{\abovecaptionskip}{0cm}
 \caption{Ablation study of the robust curriculum selection component in CurriSeg. We evaluate performance in the COD task on \textit{COD10K}. SPL and TSC are shorts for self-paced learning and teacher-student curriculum. ``w/o'' indicates without.}\label{Table:AblationCurri}
\resizebox{\columnwidth}{!}{ 
\setlength{\tabcolsep}{1.4mm}
\begin{tabular}{c|cccc|ccccc|cc|c} \toprule
\multirow{2}{*}{Metrics} & \multicolumn{4}{c|}{WCS} & \multicolumn{5}{c|}{TSSW} & \multicolumn{2}{c|}{PUE}                                  &\cellcolor{c2!20} CurriSeg       \\ \cline{2-12}
& SPL $\rightarrow$ WCS & TSC $\rightarrow$ WCS & $p_1(t)\rightarrow p(t)$ & $p_2(t)\rightarrow p(t)$ & {w/o $\omega_i^\mu$} & {w/o $\omega_i^\sigma$} &  {w/o $\omega_i^{out}$} &  {$\omega_i^{\sigma1} \rightarrow \omega_i^\sigma$} & {$\omega_i^{\sigma2} \rightarrow \omega_i^\sigma$} & w/o $\beta(t)$ & $\beta_1(t) \rightarrow \beta(t)$ &\cellcolor{c2!20} (Ours)         \\ \midrule
$M$~$\downarrow$  & \textbf{0.030}      & \textbf{0.030}             & 0.031               & 0.031               & 0.031                                  & 0.031                                     & 0.030                                    & 0.031                                                                      & 0.031                                                                      & 0.031          & 0.030                                   & \cellcolor{c2!20}\textbf{0.030} \\
$F_\beta$~$\uparrow$  & \textbf{0.738}      & 0.735                      & 0.730               & 0.726               & 0.729                                  & 0.727                                     & 0.732                                    & 0.730                                                                      & 0.727                                                                      & 0.730          & 0.732                                   &\cellcolor{c2!20} 0.736          \\
$E_\phi$~$\uparrow$   & 0.908               & \textbf{0.910}             & 0.905               & 0.902               & 0.903                                  & 0.904                                     & 0.906                                    & 0.900                                                                      & 0.903                                                                      & 0.900          & 0.907                                   &\cellcolor{c2!20} \textbf{0.910} \\
$S_\alpha$~$\uparrow$   & \textbf{0.820}      & 0.817                      & 0.815               & 0.812               & 0.814                                  & 0.810                                     & 0.810                                    & 0.808                                                                      & 0.809                                                                      & 0.814          & 0.815                                   &\cellcolor{c2!20} 0.818         \\ \bottomrule
\end{tabular}} 
\end{minipage}
\\ 
	\begin{minipage}{\textwidth}
		\centering
		\setlength{\abovecaptionskip}{0cm}
		\caption{Ablation study of anti-curriculum promotion in CurriSeg. TAL and AA are texture-aware loss and aggressive augmentation.
		}\label{Table:AblationAnti-Curri}
		\resizebox{\columnwidth}{!}{
			\setlength{\tabcolsep}{1mm}
\begin{tabular}{c|ccccccccc|c} \toprule
Metrics & Hard-only SBFT & Random SBFT & Square filter & Progressive filter & 
% Gaussian 
Blur $\rightarrow$ SBFT & 
% Additive 
Noise $\rightarrow$ SBFT & TAL $\rightarrow$ SBFT & AA $\rightarrow$ SBFT & Reverse CurriSeg &\cellcolor{c2!20} CurriSeg \\ \midrule
$M$~$\downarrow$   & 0.031          & 0.031       & 0.031         & \textbf{0.030}     & 0.033                             & 0.032                              & 0.032                                  & 0.033                                        & 0.052            &\cellcolor{c2!20} \textbf{0.030}  \\
$F_\beta$~$\uparrow$   & 0.730          & 0.728       & 0.729         & 0.734              & 0.710                             & 0.719                              & 0.725                                  & 0.718                                        & 0.632            &\cellcolor{c2!20} \textbf{0.736}  \\
$E_\phi$~$\uparrow$  & 0.899          & 0.902       & 0.905         & \textbf{0.910}     & 0.875                             & 0.866                              & 0.897                                  & 0.868                                        & 0.787            &\cellcolor{c2!20} \textbf{0.910}  \\
$S_\alpha$~$\uparrow$    & 0.813          & 0.812       & 0.814         & 0.816              & 0.804                             & 0.806                              & 0.808                                  & 0.807                                        & 0.742            &\cellcolor{c2!20} \textbf{0.818} \\ \bottomrule 
\end{tabular}}
	\end{minipage} 
        \vspace{-6mm}
\end{table*}

\begin{figure*}[t]
\begin{minipage}{1\textwidth}
\setlength{\abovecaptionskip}{0.1cm}
	\centering
\includegraphics[width=\linewidth]{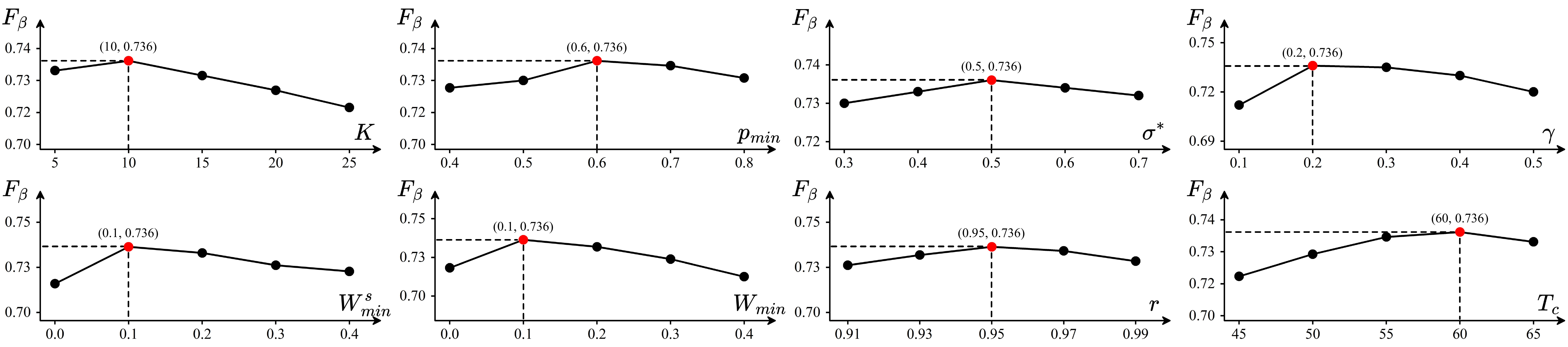}
\caption{Sensitivity analysis of hyperparameters, including $K$, $p_{min}$, $\sigma^*$, $\gamma$, $W_{min}^s$, $W_{min}$, $r$, and $T_c$.} 
% \vspace{-3mm}
\label{fig:sensitivity}
\end{minipage} \\
% \end{figure*}
% \begin{figure*}[t]
% \setlength{\abovecaptionskip}{0cm}
% 	\centering
\begin{minipage}{0.475\textwidth}
\setlength{\abovecaptionskip}{0cm}
	\centering
		\begin{subfigure}{0.235\textwidth}
\setlength{\abovecaptionskip}{0.1cm}
		\centering 
\includegraphics[width=\textwidth]{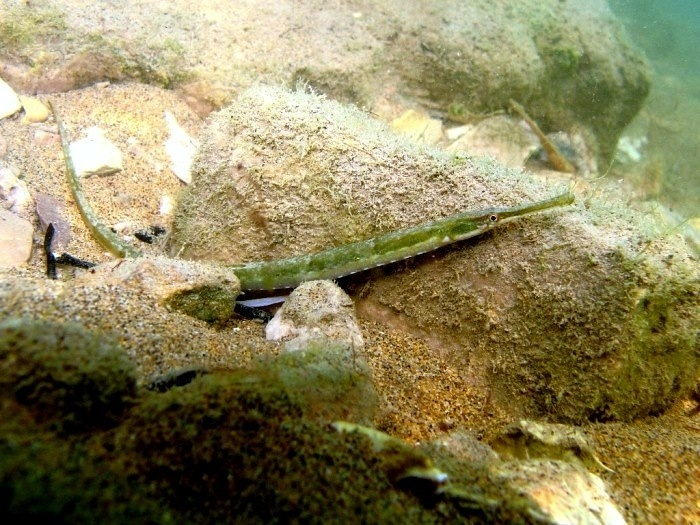}\vspace{-2pt}
		\caption{Origin}
	\end{subfigure}
	\begin{subfigure}{0.235\textwidth}  
\setlength{\abovecaptionskip}{0.1cm}
		\centering 
\includegraphics[width=\textwidth]{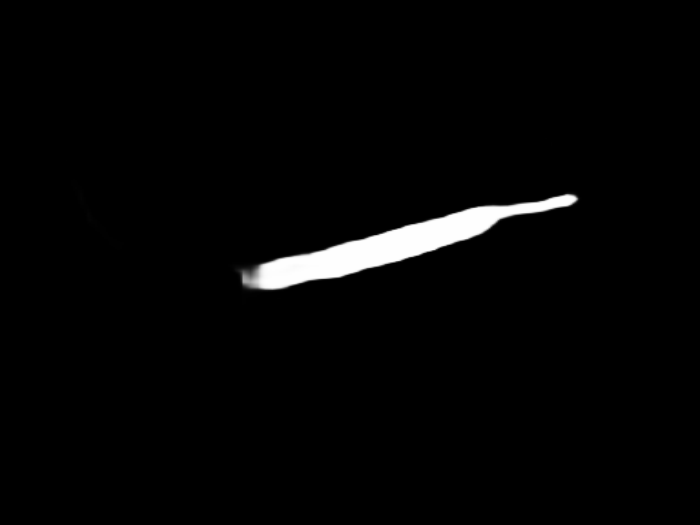}\vspace{-2pt}
		\caption{FEDER}
	\end{subfigure}
	\begin{subfigure}{0.235\textwidth}  
\setlength{\abovecaptionskip}{0.1cm}
		\centering 
\includegraphics[width=\textwidth]{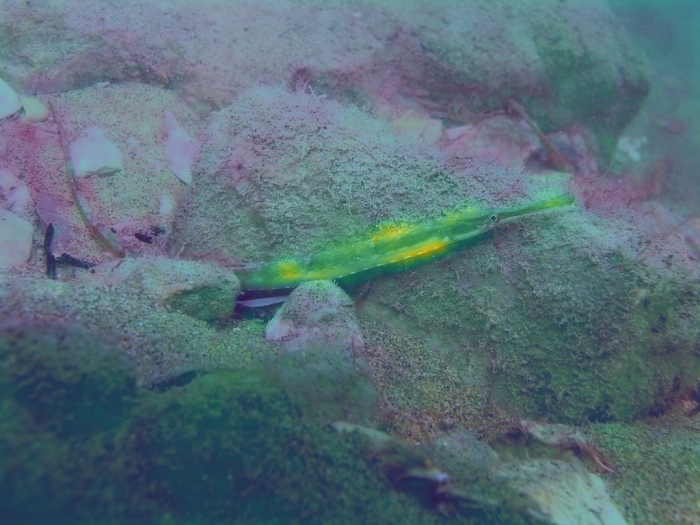}\vspace{-2pt}
\caption{Atten. map}
	\end{subfigure}
    	\begin{subfigure}{0.235\textwidth} 
\setlength{\abovecaptionskip}{0.1cm}
		\centering 
\includegraphics[width=\textwidth]{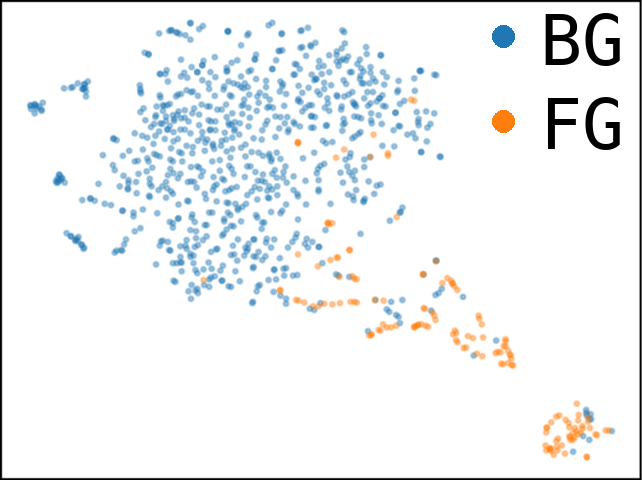}\vspace{-2pt}
\caption{t-SNE}
	\end{subfigure} \\ 
    % \vspace{0.5mm}
\begin{subfigure}{0.235\textwidth}
\setlength{\abovecaptionskip}{0cm}
		\centering
\includegraphics[width=\textwidth]{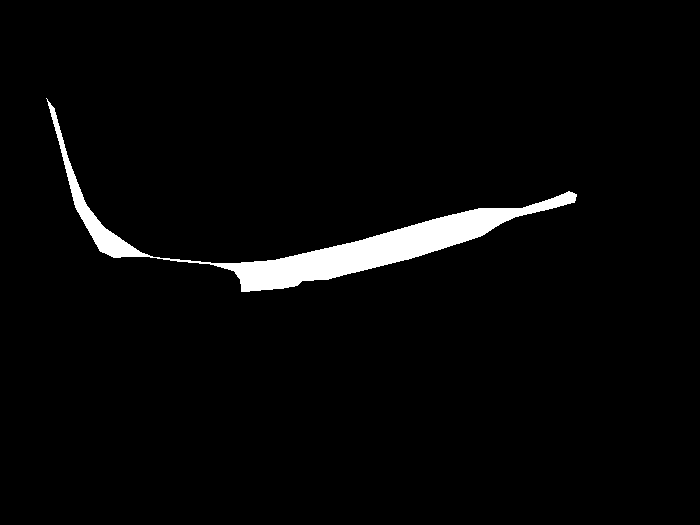} 
		\caption{GT}
	\end{subfigure}
	% \hfill
	\begin{subfigure}{0.235\textwidth} 
\setlength{\abovecaptionskip}{0cm}
		\centering 
\includegraphics[width=\textwidth]{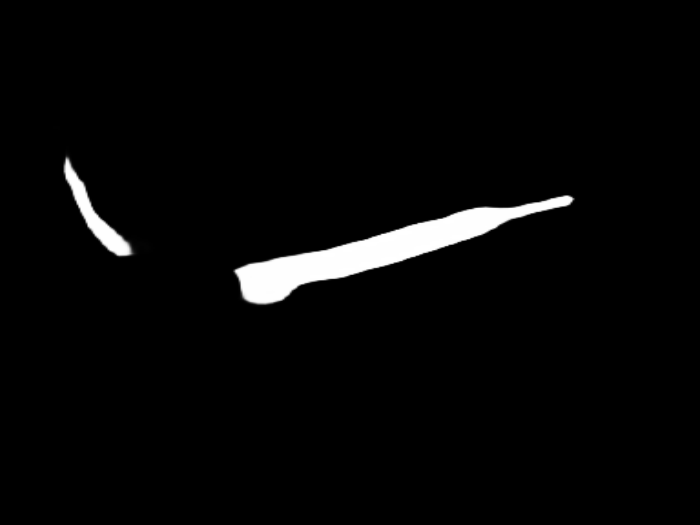}  
		\caption{FEDER+}
	\end{subfigure}
	% \hfill
	\begin{subfigure}{0.235\textwidth}  
\setlength{\abovecaptionskip}{0cm}
		\centering 
\includegraphics[width=\textwidth]{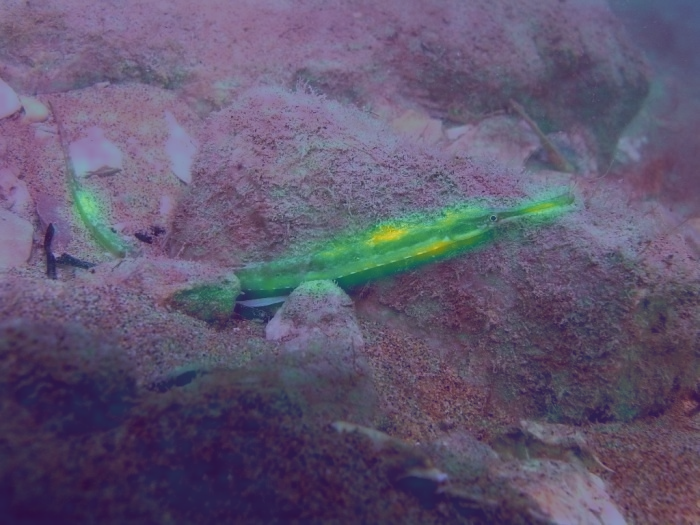} 
		\caption{Atten. map}
	\end{subfigure}
    % \hfill
    	\begin{subfigure}{0.235\textwidth} 
\setlength{\abovecaptionskip}{0cm}
		\centering 
\includegraphics[width=\textwidth]{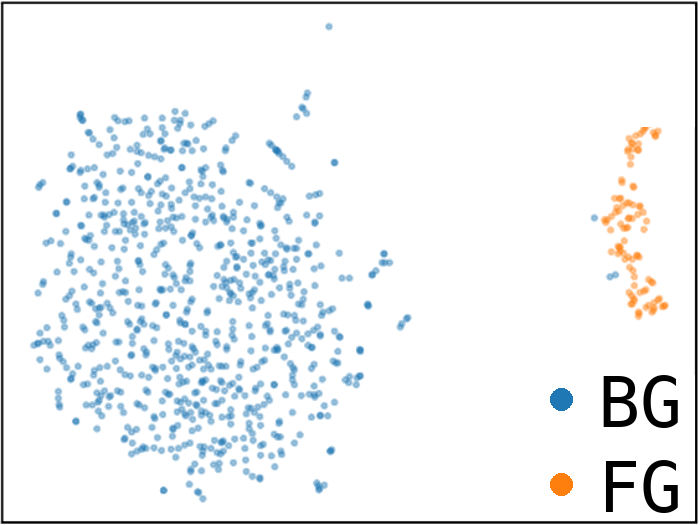} 
		\caption{t-SNE}
	\end{subfigure}
\caption{Visualization of attention map and t-SNE of foreground ({\color{orange}orange}) and background ({\color{blue}blue}) features. (b)-(d) and (f)-(h) correspond to the prediction, attention map, and t-sne of FEDER trained under the standard setting and our CurriSeg framework.}
\label{fig:tsne}
\end{minipage}\hspace{4mm}
% \vspace{-3mm}
% \end{figure}
% \begin{figure}[t]
% \setlength{\abovecaptionskip}{0.05cm}
% 	\centering
\begin{minipage}{0.475\textwidth}
\setlength{\abovecaptionskip}{0cm}
	\centering
		\begin{subfigure}{0.49\textwidth}
\setlength{\abovecaptionskip}{0.1cm}
		\centering 
\includegraphics[width=\textwidth]{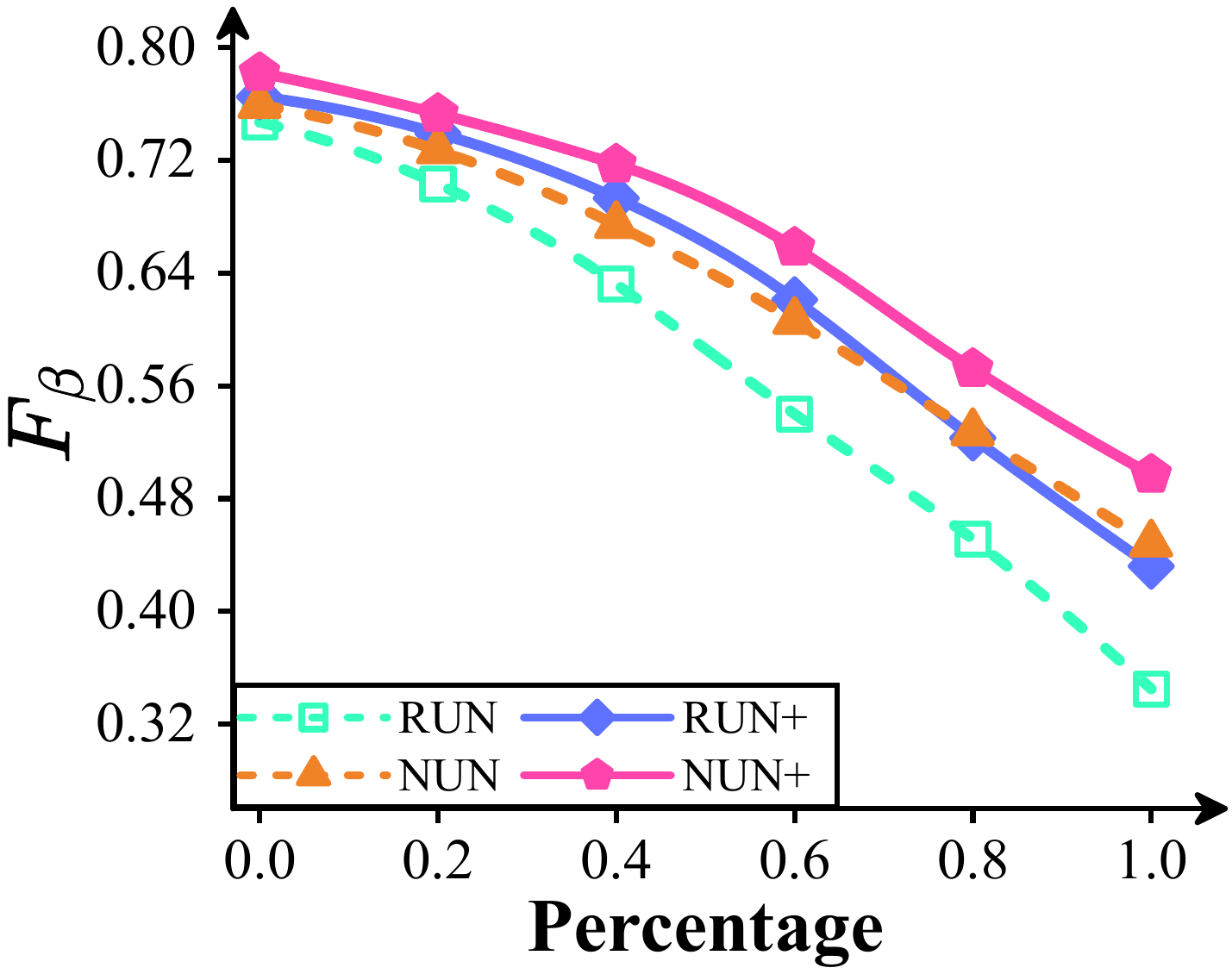}\vspace{-2pt}
	\end{subfigure}
	\begin{subfigure}{0.49\textwidth}  
\setlength{\abovecaptionskip}{0.1cm}
		\centering 
\includegraphics[width=\textwidth]{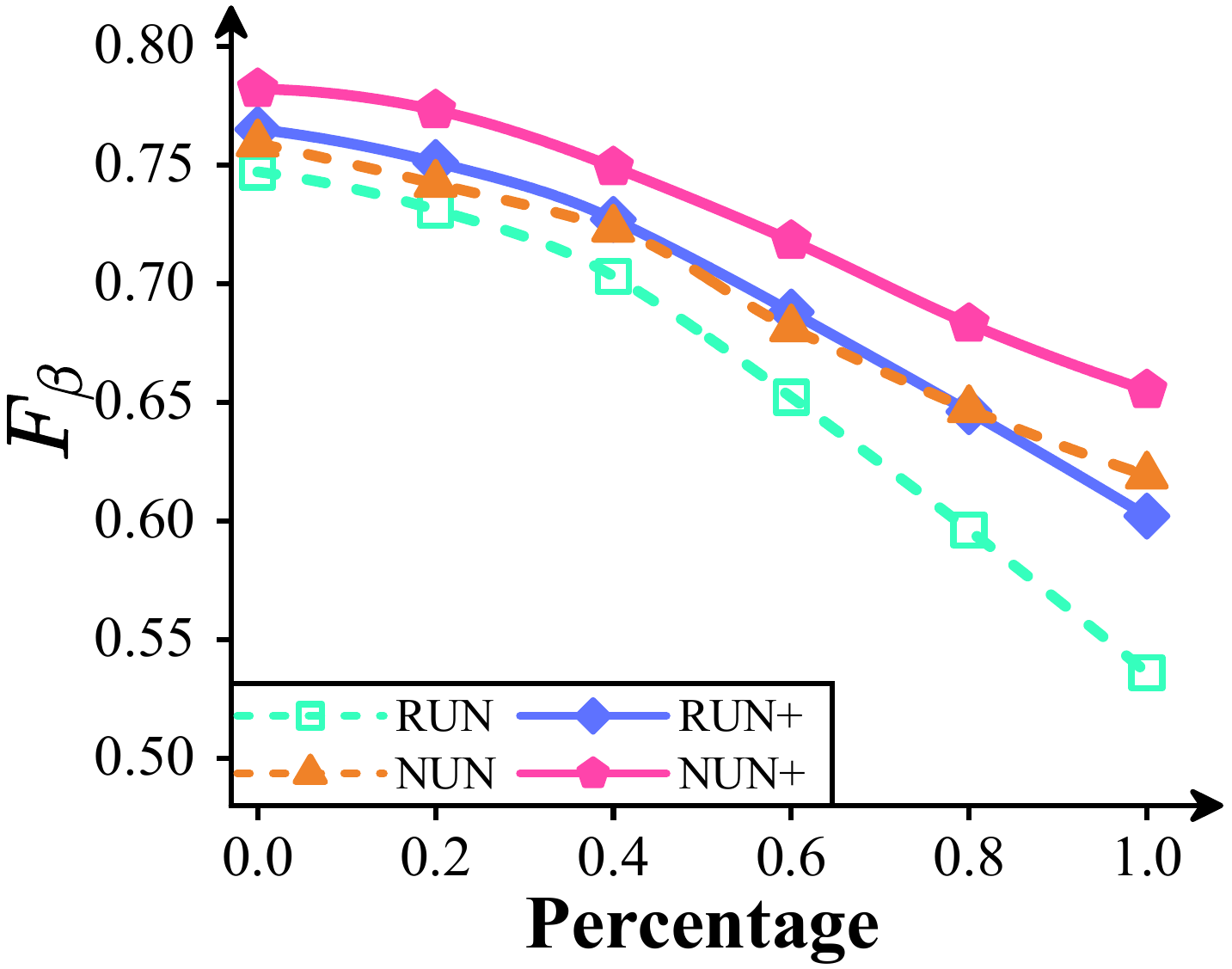}\vspace{-2pt}
	\end{subfigure}
\caption{Robustness under increasing training data degradation. \textbf{Left}: High-frequency information loss (blur, low-light, haze). \textbf{Right}: High-frequency interference (noise). 
CurriSeg maintains superior performance across all degradation ratios, indicating the potential of our framework in real-world degradation scenarios.
}
\label{fig:degradation}
\end{minipage}
\vspace{-3mm}
\end{figure*}
\begin{table*}[t]
\centering
\begin{minipage}[c]{1\textwidth}
\centering
\setlength{\abovecaptionskip}{0.3cm}
\caption{Generalization of CurriSeg on other settings, including weak supervision with scribble on \textit{COD10K} (WS-SAM~\cite{he2023weaklysupervised} and SEE~\cite{he2025segment}), semi-supervision with 1/16 labeled data on \textit{COD10K} (CoSOD~\cite{chakraborty2024unsupervised} and SEE), multi-modality with depth on \textit{COD10K} (DSAM~\cite{yu2024exploring} and MultiCOS~\cite{fang2025integrating}), and video segmentation on \textit{CAD} \cite{cheng2022implicit} (STL-Net~\cite{cheng2022implicit} and ZoomNext~\cite{pang2024zoomnext}). Metrics are reported according to each setting.
% The metrics are selected in the corresponding setting.
}
		\label{table:OtherSetting}
        \vspace{-3mm}
\begin{subtable}{0.232\textwidth}
\centering
\setlength{\abovecaptionskip}{0cm}
\caption{Weak supervision.}
		\label{table:OtherSettingWeak}
\resizebox{\columnwidth}{!}{
\setlength{\tabcolsep}{1mm}
\begin{tabular}{l|cccc} \toprule
Methods & {\cellcolor{gray!40}$M$~$\downarrow$}                                  & {\cellcolor{gray!40}$F_\beta$~$\uparrow$}                               & {\cellcolor{gray!40}$E_\phi$~$\uparrow$}                               & \multicolumn{1}{c}{\cellcolor{gray!40}$S_\alpha$~$\uparrow$} \\ \midrule
WS-SAM
% ~\cite{he2023weaklysupervised}
& 0.038 & 0.719 & 0.878 & 0.803 \\
\rowcolor{c2!20} WS-SAM+ & 0.035 & 0.733 & 0.886 & 0.813 \\
SEE
% ~\cite{he2025segment}     
& 0.036 & 0.729 & 0.883 & 0.807 \\
\rowcolor{c2!20} SEE+    & \textbf{0.033} & \textbf{0.738} & \textbf{0.893} & \textbf{0.816} \\ \bottomrule
\end{tabular}}
\end{subtable} 
\begin{subtable}{0.221\textwidth}
\centering
\setlength{\abovecaptionskip}{0cm}
\caption{Semi-supervision.}
		\label{table:OtherSettingSemi}
\resizebox{\columnwidth}{!}{
\setlength{\tabcolsep}{1mm}
\begin{tabular}{l|cccc} \toprule
Methods & {\cellcolor{gray!40}$M$~$\downarrow$}                                  & {\cellcolor{gray!40}$F_\beta$~$\uparrow$}                               & {\cellcolor{gray!40}$E_\phi$~$\uparrow$}                               & \multicolumn{1}{c}{\cellcolor{gray!40}$S_\alpha$~$\uparrow$} \\ \midrule
CoSOD   & 0.055 & 0.650 & 0.795 & 0.740 \\
\rowcolor{c2!20} CoSOD+  & 0.051 & 0.672 & 0.810 & 0.749 \\
SEE     & 0.046 & 0.679 & 0.803 & 0.745 \\
\rowcolor{c2!20} SEE+    & \textbf{0.043} & \textbf{0.695} & \textbf{0.817} & \textbf{0.757} \\ \bottomrule
\end{tabular}}
\end{subtable}
\begin{subtable}{0.258\textwidth}
\centering
\setlength{\abovecaptionskip}{0cm}
\caption{Multi-modality.}
		\label{table:OtherSettingMulti}
\resizebox{\columnwidth}{!}{
\setlength{\tabcolsep}{1mm}
\begin{tabular}{l|cccc} \toprule
Methods & {\cellcolor{gray!40}$M$~$\downarrow$}                                  & {\cellcolor{gray!40}$F_\beta^{max}$~$\uparrow$}                               & {\cellcolor{gray!40}$E_\phi^{max}$~$\uparrow$}                               & \multicolumn{1}{c}{\cellcolor{gray!40}$S_\alpha$~$\uparrow$} \\ \midrule
DSAM      & 0.033 & 0.807                      & 0.931                      & 0.846 \\
\rowcolor{c2!20} DSAM+     & 0.030 & 0.821                      & 0.942                      & 0.853 \\
MultiCOS  & 0.020 & 0.950                      & 0.946                      & 0.880 \\
\rowcolor{c2!20} MultiCOS+ & \textbf{0.018} & \textbf{0.958}                     & \textbf{0.953}                      & \textbf{0.883} \\ \bottomrule
\end{tabular}}
\end{subtable}
\begin{subtable}{0.247\textwidth}
\centering
\setlength{\abovecaptionskip}{0cm}
\caption{Video segmentation.}
		\label{table:OtherSettingVideo}
\resizebox{\columnwidth}{!}{
\setlength{\tabcolsep}{1mm}
\begin{tabular}{l|cccc} \toprule
Methods & {\cellcolor{gray!40}$M$~$\downarrow$}                                  & {\cellcolor{gray!40}$F_\beta^{w}$~$\uparrow$}                               & {\cellcolor{gray!40}$E_\phi^{max}$~$\uparrow$}                               & \multicolumn{1}{c}{\cellcolor{gray!40}$S_\alpha$~$\uparrow$} \\ \midrule
STL-Net   & 0.030 & 0.481                & 0.845                      & 0.696 \\
\rowcolor{c2!20} STL-Net+  & 0.026 & 0.505                & 0.861                      & 0.706 \\
ZoomNext  & 0.020 & 0.593                & 0.865                      & 0.757 \\
\rowcolor{c2!20} ZoomNext+ & \textbf{0.018} & \textbf{0.611}                & \textbf{0.873}                      & \textbf{0.761} \\ \bottomrule
\end{tabular}}
\end{subtable}
\end{minipage}
\vspace{-6mm}
\end{table*}

\section{Experiments}
% \vspace{-0.1cm}
\noindent \textbf{Experimental setup}. Our CurriSeg framework is implemented in PyTorch and trained on two NVIDIA RTX 4090 GPUs, using the Adam optimizer with momentum terms parameters $(\beta_1,\beta_2)=(0.9,0.999)$. We adopt an initial warm-up period of 10 epochs in which all training samples are used without curriculum filtering, allowing the model to acquire a basic representation capacity before difficulty-based selection is applied.
% The selection percentile $p(t)$ is initialized at $60\%$ and increased by $10\%$ per 10 epochs. 
Unless specified, the curriculum phase length is set to $T_c = 60$ epochs, followed by an anti-curriculum fine-tuning phase up to $T=70$ epochs. 
When integrating CurriSeg with existing CECS backbones, all input images are resized to $352\times 352$ during both training and testing, and all other hyperparameters follow the original configurations of the corresponding backbone methods.

% \vspace{-0.2cm}
\subsection{Comparative Evaluation}
% \vspace{-0.1cm}
We conduct extensive experiments across CECS tasks, with datasets and metrics detailed in~\cref{Sec:Dataset} in the appendix. All results follow standardized evaluation protocols.
% We perform extensive experiments across several CECS tasks. The datasets and metrics are described in~\cref{Sec:Dataset}. To ensure fairness and reproducibility, all reported results are obtained using standardized evaluation protocols.

\noindent \textbf{Camouflaged object detection}.
As shown in \cref{table:CODQuanti}, integrating our CurriSeg into cutting-edge methods (denoted by ``+'') consistently improves performance across ResNet50~\cite{he2016deep}, Res2Net50~\cite{gao2019res2net}, and PVT V2~\cite{wang2022pvt}.
Moreover, the qualitative results in \cref{fig:visualization} demonstrate that models optimized under our CurriSeg yield more accurate segmentation, even in highly challenging scenarios. Together, these results substantiate the effectiveness of our dual-phase curriculum strategy. 
Beyond performance gains, CurriSeg also improves training efficiency. As reported in \cref{table:overhead}, training time is reduced by 28.9\% -48.4\% across backbones, since curriculum selection concentrates computation on informative samples rather than all data. GPU memory overhead is negligible ($\leq$0.1G), since no extra parameters are introduced. 
Hence, our CurriSeg is more efficient than the standard manner.

\begin{figure}[t]
\begin{minipage}{1\linewidth}
\centering
\setlength{\abovecaptionskip}{0cm}
\includegraphics[width=\linewidth]{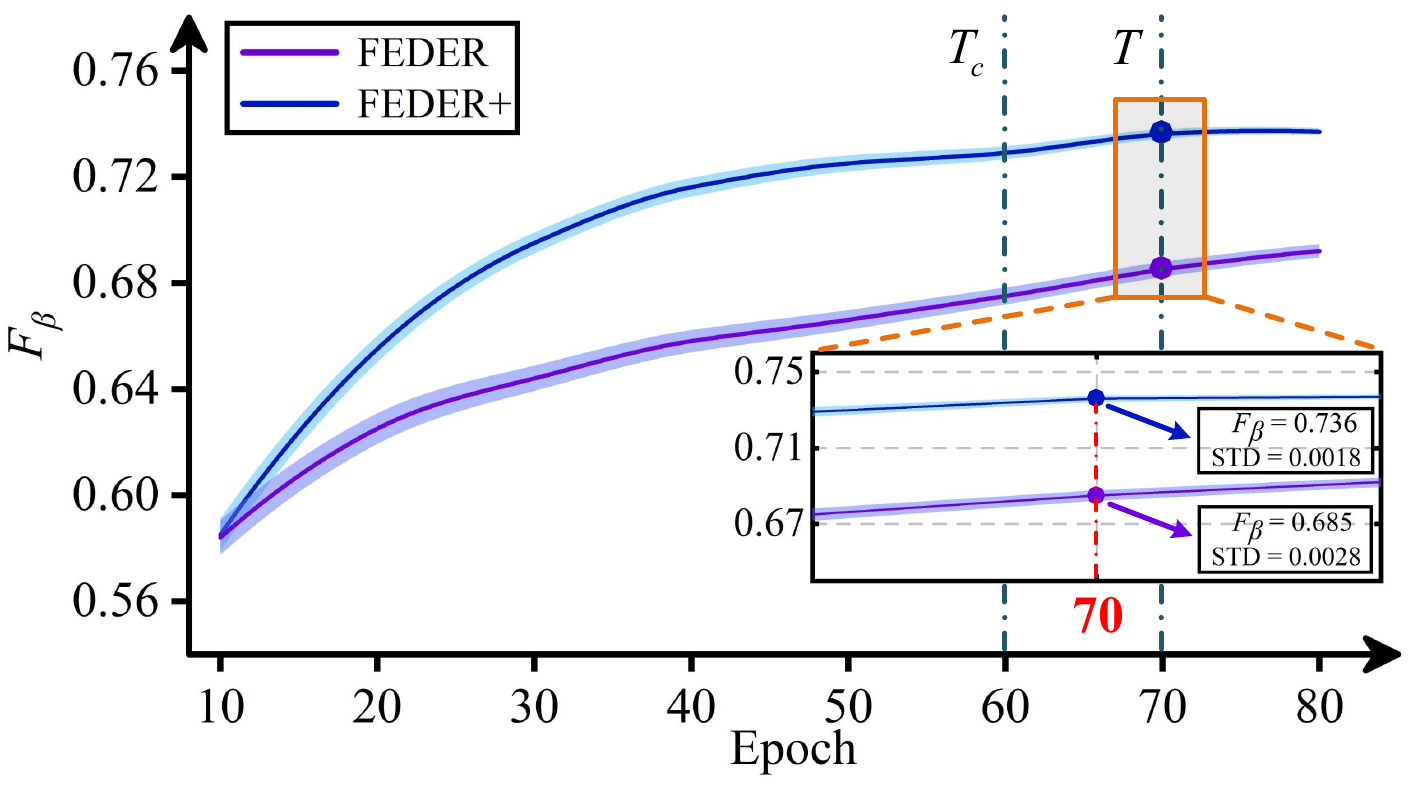}
\caption{Performance ($F_\beta$) vs.\ training epoch on COD10K. Phase 1: 0-$T_c$ and Phase 2: $T_c$-$T$. The initial 10 epochs are used for warm-up, where FEDER shares the same training details with FEDER+, and thus be omitted for clarification. 
% CurriSeg achieves higher performance with reduced variance.
}
\label{fig:epoch_curve}
\end{minipage}
\vspace{-1.5mm}
\end{figure}

\noindent \textbf{Polyp image segmentation}.
Experiments are conducted on the PIS task using \textit{CVC-ColonDB} and \textit{ETIS}. In line with RUN~\cite{he2025run}, PVT V2 is employed as the backbone. As evidenced by the quantitative results in \cref{table:PISQuanti} and the visualizations in \cref{fig:visualization}, models trained under our CurriSeg consistently outperform their original versions.

\noindent \textbf{Transparent object detection}.
As reported in \cref{table:TODQuanti,fig:visualization}, CurriSeg integration yields consistent gains over existing methods, highlighting its utility as a general performance enhancer for perception in autonomous driving.

\noindent \textbf{Concealed defect detection}.
We validate the generalizability of models trained with CurriSeg on CDD. Models pre-trained on COD are directly transferred to the \textit{CDS2K} dataset for defect segmentation, following RUN~\cite{he2025run}. As shown in \cref{table:CDDQuanti,fig:visualization}, models optimized under CurriSeg consistently achieve superior performance, demonstrating enhanced cross-domain generalization.

\begin{table*}[t]
\centering
\begin{minipage}[c]{1\textwidth}
\centering
\setlength{\abovecaptionskip}{0.3cm}
\caption{Generalization of CurriSeg on other dense prediction tasks with cutting-edge methods, including semantic segmentation on \textit{ADE20K}~\cite{zhou2017scene} (PEM~\cite{cavagnero2024pem} and CGRSeg~\cite{ni2024context}), instance segmentation on \textit{COCO}~\cite{lin2014microsoft} (M2Form and FasIns are short for Mask2Former~\cite{cheng2022masked} and FastInst~\cite{he2023fastinst}), infrared small target detection on \textit{IRSTD-1k}~\cite{zhang2022isnet} (ISNet~\cite{zhang2022isnet} and IRSAM~\cite{zhang2024irsam}), and shadow detection on \textit{SBU}~\cite{vicente2016large} (SARA~\cite{sun2023adaptive} and spider~\cite{zhao2024spider}), with BER~\cite{vicente2015leave} used for evaluation.}
		\label{table:OtherTasks}
        \vspace{-3mm}
\begin{subtable}{0.265\textwidth}
\centering
\setlength{\abovecaptionskip}{0cm}
\caption{Semantic segmentation.}
		\label{table:OtherTasksSemantic}
\resizebox{\columnwidth}{!}{
\setlength{\tabcolsep}{0.8mm}
\begin{tabular}{l|cc|cc} \toprule
Metric & PEM                      & \cellcolor{c2!20} PEM+                     & CGRSeg                   & \cellcolor{c2!20} CGRSeg+                  \\ \hline
IoU $\uparrow$   & \multicolumn{1}{c}{45.0} & \cellcolor{c2!20} {45.9} & \multicolumn{1}{c}{45.5} & \cellcolor{c2!20} {\textbf{46.8}} \\\bottomrule
\end{tabular}}
\end{subtable} 
\begin{subtable}{0.24\textwidth}
\centering
\setlength{\abovecaptionskip}{0cm}
\caption{Instance segmentation.}
		\label{table:OtherTasksInstance}
\resizebox{\columnwidth}{!}{
\setlength{\tabcolsep}{0.6mm}
\begin{tabular}{l|cc|cc} \toprule
Metric & M2Form &\cellcolor{c2!20} M2Form+ & FasIns &\cellcolor{c2!20} FasIns+ \\ \midrule
AP $\uparrow$     & 38.0        &\cellcolor{c2!20} 38.6         & 38.6     &\cellcolor{c2!20} \textbf{39.1}              \\ \bottomrule
\end{tabular}}
\end{subtable}
\begin{subtable}{0.243\textwidth}
\centering
\setlength{\abovecaptionskip}{0cm}
\caption{Infrared small target detection.}
		\label{table:OtherTasksSmallInfrared}
\resizebox{\columnwidth}{!}{
\setlength{\tabcolsep}{1.1mm}
\begin{tabular}{l|cc|cc} \toprule
Metric & ISNet &\cellcolor{c2!20}  ISNet+ & IRSAM &\cellcolor{c2!20}  IRSAM+ \\ \midrule
IoU $\uparrow$    & 68.77 &\cellcolor{c2!20}  70.62  & 73.69 &\cellcolor{c2!20}  \textbf{75.52}  \\ \bottomrule
\end{tabular}}
\end{subtable}
\begin{subtable}{0.236\textwidth}
\centering
\setlength{\abovecaptionskip}{0cm}
\caption{Shadow detection.}
		\label{table:OtherTasksShadow}
\resizebox{\columnwidth}{!}{
\setlength{\tabcolsep}{1mm}
\begin{tabular}{l|cc|cc} \toprule
Metric & SARA  &\cellcolor{c2!20}  SARA+ & Spider &\cellcolor{c2!20}  Spider+ \\ \midrule
BER  $\downarrow$   & 0.043 &\cellcolor{c2!20}  0.040 & 0.040  &\cellcolor{c2!20}  \textbf{0.038}   \\ \bottomrule
\end{tabular}}
\end{subtable}
\end{minipage}
\vspace{-3mm}
\end{table*}

\begin{table*}[t]
\centering
\begin{minipage}[c]{1\textwidth}
\centering
\setlength{\abovecaptionskip}{0.3cm}
\caption{Compatibility of our CurriSeg with advanced architectures (attention-based FSEL~\cite{sun2025frequency}, multiscale-based ZoomNet~\cite{pang2022zoom}, and uncertainty-based UGTR~\cite{yang2021uncertainty}) and foundation models (SAM-adapter (SAM-a)~\cite{chen2023sam}, SAM2-adapter (SAM2-a)~\cite{chen2024sam2}, and SAM3-adapter (SAM3-a)~\cite{chen2025sam3}) on \textit{COD10K}. In (b)-(d), the suffix ``*'' means modifying the general CurriSeg framework with simple, architecture-specific changes. The further performance improvement indicates the potential of our CurriSeg in facilitating the community.
}
		\label{table:Compatibility}
        \vspace{-3mm}
\begin{subtable}{0.0487\textwidth}
\centering
\setlength{\abovecaptionskip}{0cm}
\caption{}
\resizebox{\columnwidth}{!}{
\setlength{\tabcolsep}{1mm}
\begin{tabular}{l|} \toprule
Metr. \\ \midrule
{$M$~$\downarrow$} \\
{$F_\beta$~$\uparrow$} \\
{$E_\phi$~$\uparrow$}  \\
{$S_\alpha$~$\uparrow$} \\ \bottomrule
\end{tabular}}
\end{subtable}
\begin{subtable}{0.154\textwidth}
\centering
\setlength{\abovecaptionskip}{0cm}
\caption{Attention.}
		\label{table:CompatibilityAttention}
\resizebox{\columnwidth}{!}{
\setlength{\tabcolsep}{1mm}
\begin{tabular}{ccc} \toprule
FSEL  & \cellcolor{c2!20}FSEL+ & \cellcolor{c2!20}FSEL* \\ \midrule
0.032 & \cellcolor{c2!20}0.030  & \cellcolor{c2!20}\textbf{0.029}   \\
0.722 & \cellcolor{c2!20}0.742  & \cellcolor{c2!20}\textbf{0.749}   \\
0.891 & \cellcolor{c2!20}0.909  & \cellcolor{c2!20}\textbf{0.916}   \\
0.822 & \cellcolor{c2!20}0.831  & \cellcolor{c2!20}\textbf{0.834}  \\ \bottomrule
\end{tabular}}
\end{subtable} 
\begin{subtable}{0.221\textwidth}
\centering
\setlength{\abovecaptionskip}{0cm}
\caption{Multiscale.}
		\label{table:CompatibilityMultiscale}
\resizebox{\columnwidth}{!}{
\setlength{\tabcolsep}{1mm}
\begin{tabular}{ccc} \toprule
ZoomNet & \cellcolor{c2!20}ZoomNet+ & \cellcolor{c2!20}ZoomNet* \\ \midrule
0.029   & \cellcolor{c2!20}0.028  & \cellcolor{c2!20}\textbf{0.027}   \\
0.740   & \cellcolor{c2!20}0.753  & \cellcolor{c2!20}\textbf{0.758}   \\
0.888   & \cellcolor{c2!20}0.898  & \cellcolor{c2!20}\textbf{0.907}   \\
0.838   & \cellcolor{c2!20}0.842  & \cellcolor{c2!20}\textbf{0.845}  \\ \bottomrule
\end{tabular}}
\end{subtable}
\begin{subtable}{0.172\textwidth}
\centering
\setlength{\abovecaptionskip}{0cm}
\caption{Uncertainty.}
		\label{table:CompatibilityUncertainty}
\resizebox{\columnwidth}{!}{
\setlength{\tabcolsep}{1mm}
\begin{tabular}{ccc} \toprule
UGTR  & \cellcolor{c2!20}UGTR+ & \cellcolor{c2!20}UGTR* \\ \midrule
0.036 & \cellcolor{c2!20}0.034  & \cellcolor{c2!20}\textbf{0.032}   \\
0.670 & \cellcolor{c2!20}0.688  & \cellcolor{c2!20}\textbf{0.706}   \\
0.852 & \cellcolor{c2!20}0.865  & \cellcolor{c2!20}\textbf{0.889}   \\
0.817 & \cellcolor{c2!20}0.825  & \cellcolor{c2!20}\textbf{0.830}  \\ \bottomrule
\end{tabular}}
\end{subtable}
\begin{subtable}{0.383\textwidth}
\centering
\setlength{\abovecaptionskip}{0cm}
\caption{Foundation model adapters.}
		\label{table:CompatibilityFoundation}
\resizebox{\columnwidth}{!}{
\setlength{\tabcolsep}{1mm}
\begin{tabular}{cc|cc|cc} \toprule
% SAM-adapter
SAM-a & \cellcolor{c2!20}SAM-a+ & SAM2-a & \cellcolor{c2!20}SAM2-a+ & SAM3-a & \cellcolor{c2!20}SAM3-a+ \\ \midrule
0.025       &\cellcolor{c2!20} 0.023        & 0.018        &\cellcolor{c2!20} 0.016         & 0.015        &\cellcolor{c2!20} \textbf{0.014}         \\
0.800       &\cellcolor{c2!20} 0.822        & 0.848        &\cellcolor{c2!20} 0.869         & 0.883        &\cellcolor{c2!20} \textbf{0.892}         \\
0.918       &\cellcolor{c2!20} 0.935        & 0.950        &\cellcolor{c2!20} 0.962         & 0.965        &\cellcolor{c2!20} \textbf{0.968}         \\
0.883       &\cellcolor{c2!20} 0.888        & 0.899        &\cellcolor{c2!20} 0.903         & 0.927        &\cellcolor{c2!20} \textbf{0.930}   \\ \bottomrule     
\end{tabular}}
\end{subtable}
\end{minipage} \\ \vspace{0.5mm}
% \vspace{-4mm}
% \end{table*}
% \begin{figure*}[t]
\begin{minipage}[c]{1\textwidth}
\setlength{\abovecaptionskip}{0cm}
\centering
		\begin{subfigure}{0.12\textwidth}
\setlength{\abovecaptionskip}{0.1cm}
		\centering 
\includegraphics[width=\textwidth]{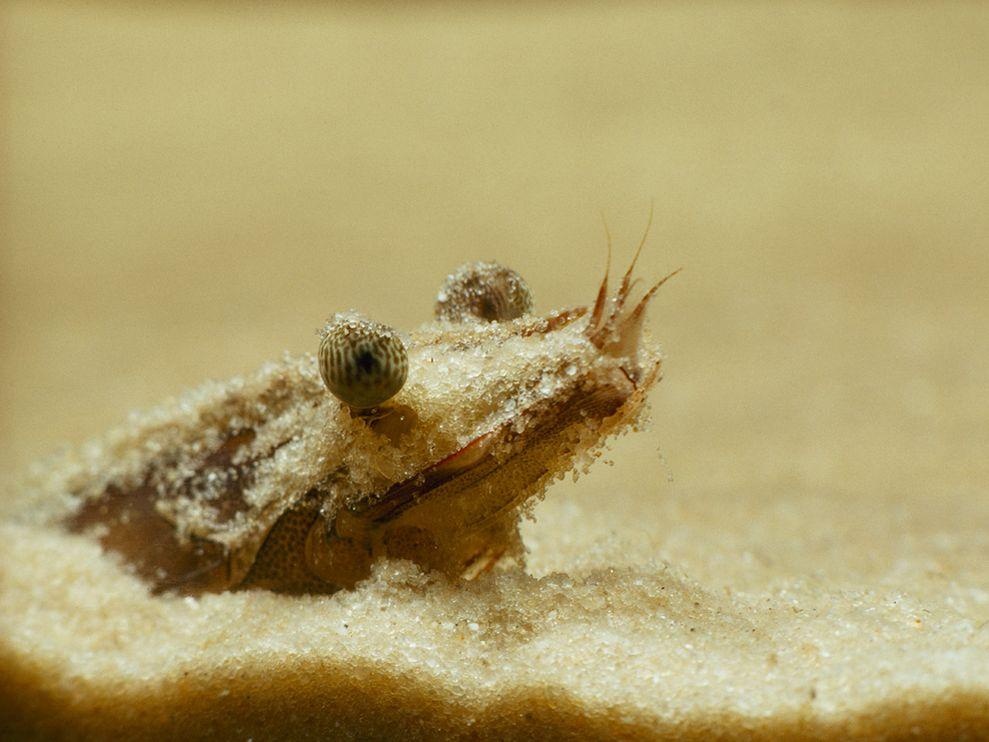}\vspace{-2pt}
%		\caption{Origin}
	\end{subfigure}
	\begin{subfigure}{0.12\textwidth}  
\setlength{\abovecaptionskip}{0.1cm}
		\centering 
\includegraphics[width=\textwidth]{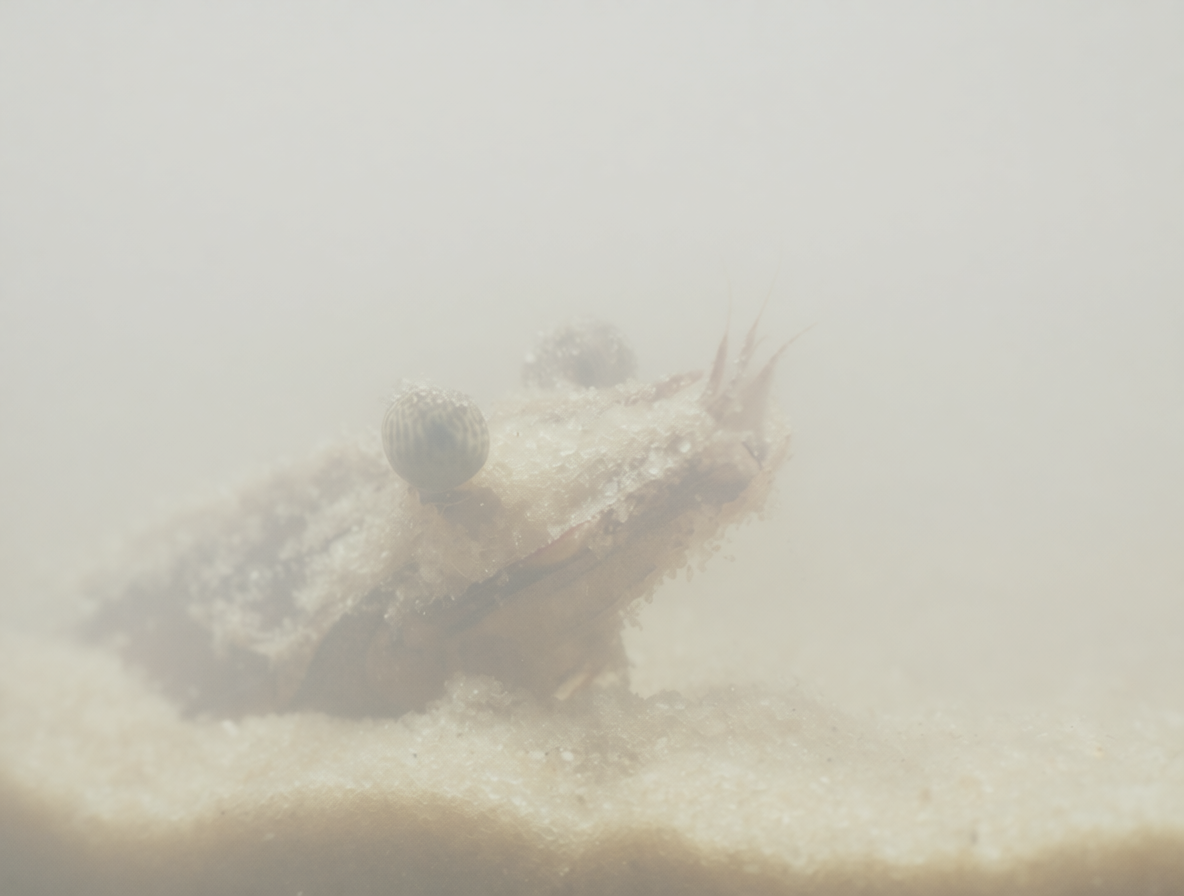}\vspace{-2pt}
%		\caption{Degradation}
	\end{subfigure}
	\begin{subfigure}{0.12\textwidth}  
\setlength{\abovecaptionskip}{0.1cm}
		\centering 
\includegraphics[width=\textwidth]{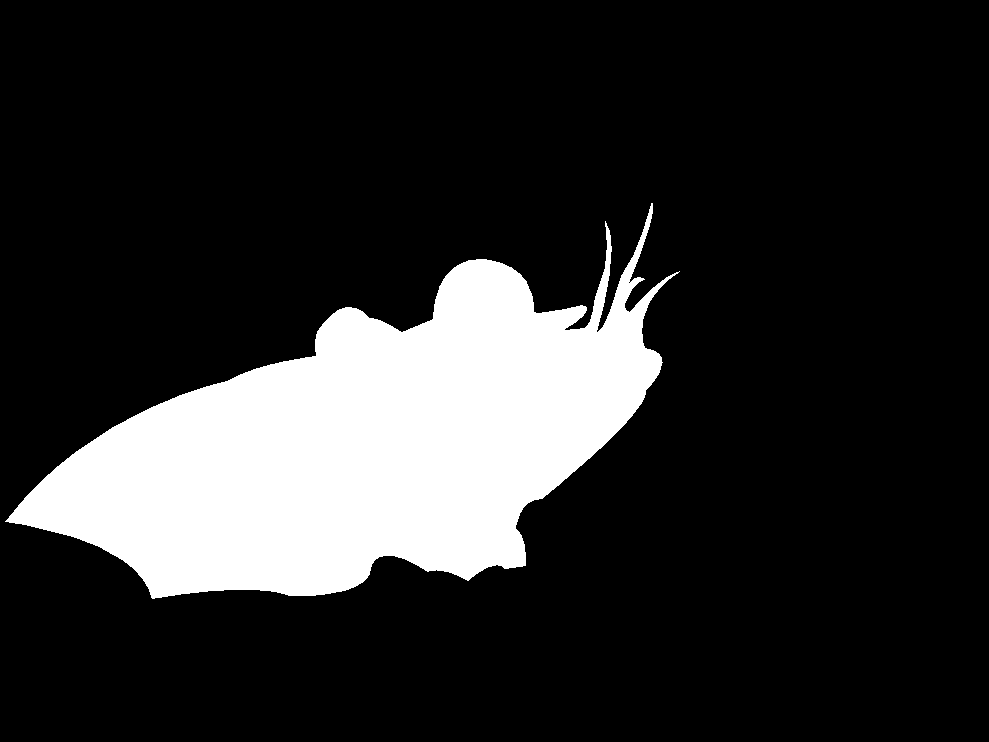}\vspace{-2pt}
%     \caption{GT}
	\end{subfigure}
    	\begin{subfigure}{0.12\textwidth} 
\setlength{\abovecaptionskip}{0.1cm}
		\centering 
\includegraphics[width=\textwidth]{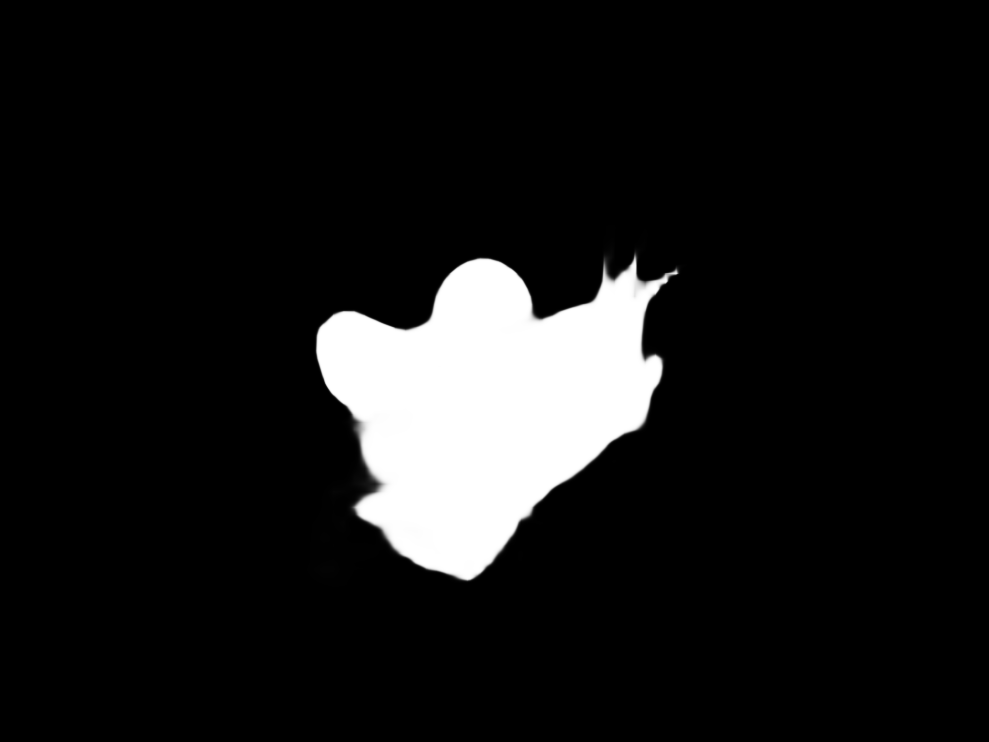}\vspace{-2pt}
%   \caption{Initial}
	\end{subfigure} 
		\begin{subfigure}{0.12\textwidth}
\setlength{\abovecaptionskip}{0.1cm}
		\centering 
\includegraphics[width=\textwidth]{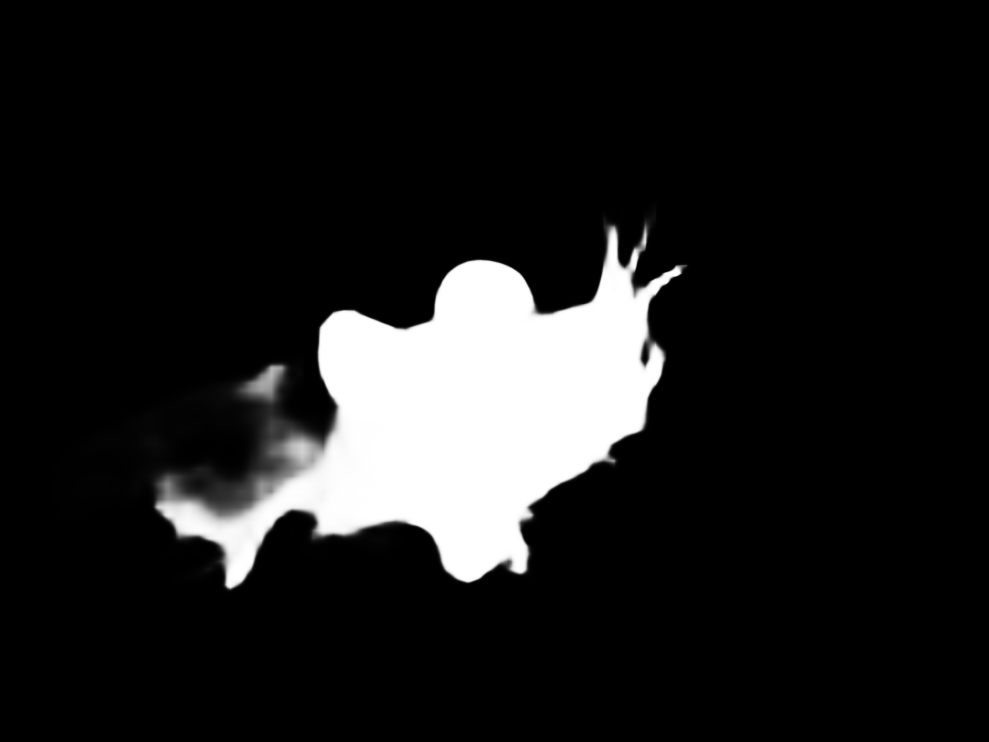}\vspace{-2pt}
%		\caption{Mid}
	\end{subfigure}
	\begin{subfigure}{0.12\textwidth}  
\setlength{\abovecaptionskip}{0.1cm}
		\centering 
\includegraphics[width=\textwidth]{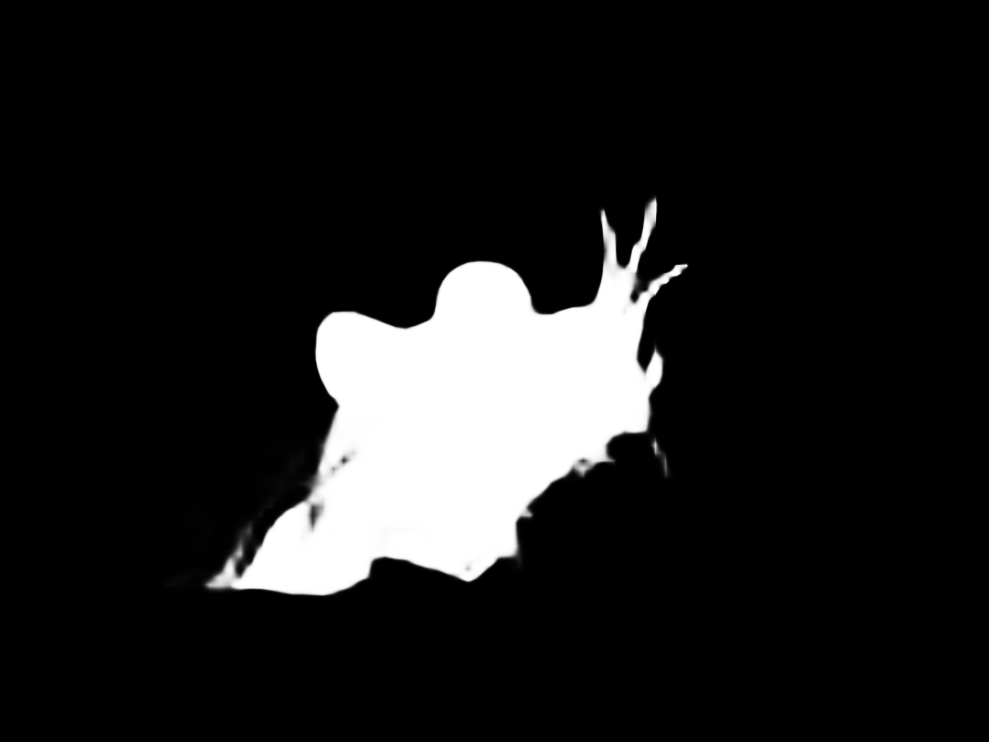}\vspace{-2pt}
%		\caption{Final}
	\end{subfigure}
	\begin{subfigure}{0.12\textwidth}  
\setlength{\abovecaptionskip}{0.1cm}
		\centering 
\includegraphics[width=\textwidth]{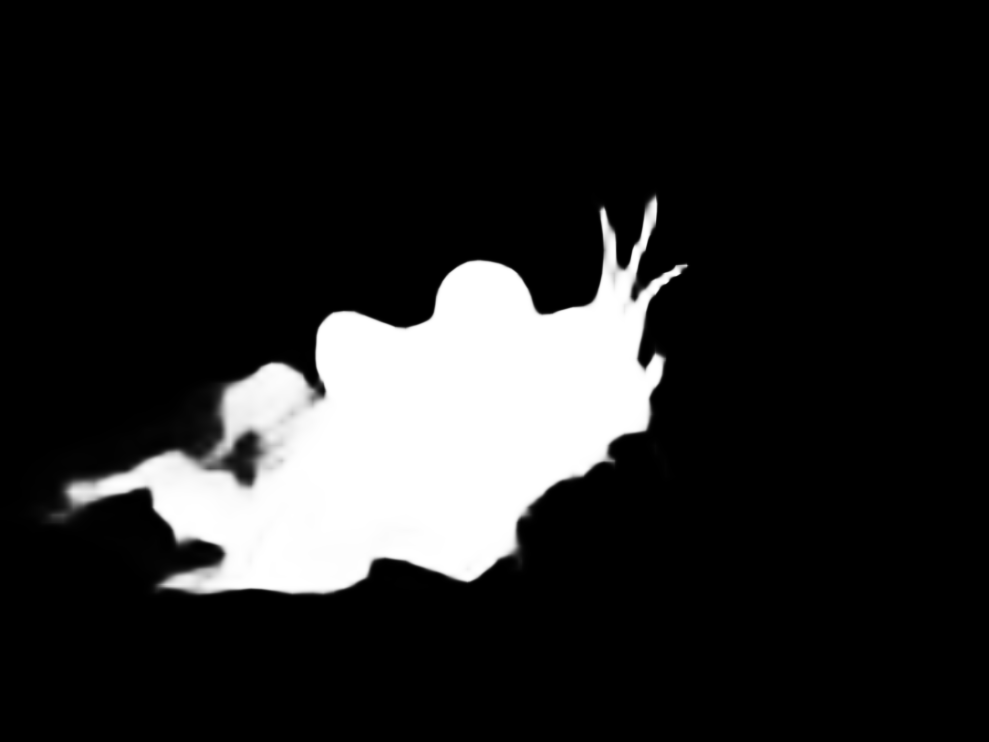}\vspace{-2pt}
%     \caption{+ Anti-Curri}
	\end{subfigure}
    	\begin{subfigure}{0.12\textwidth} 
\setlength{\abovecaptionskip}{0.1cm}
		\centering 
\includegraphics[width=\textwidth]{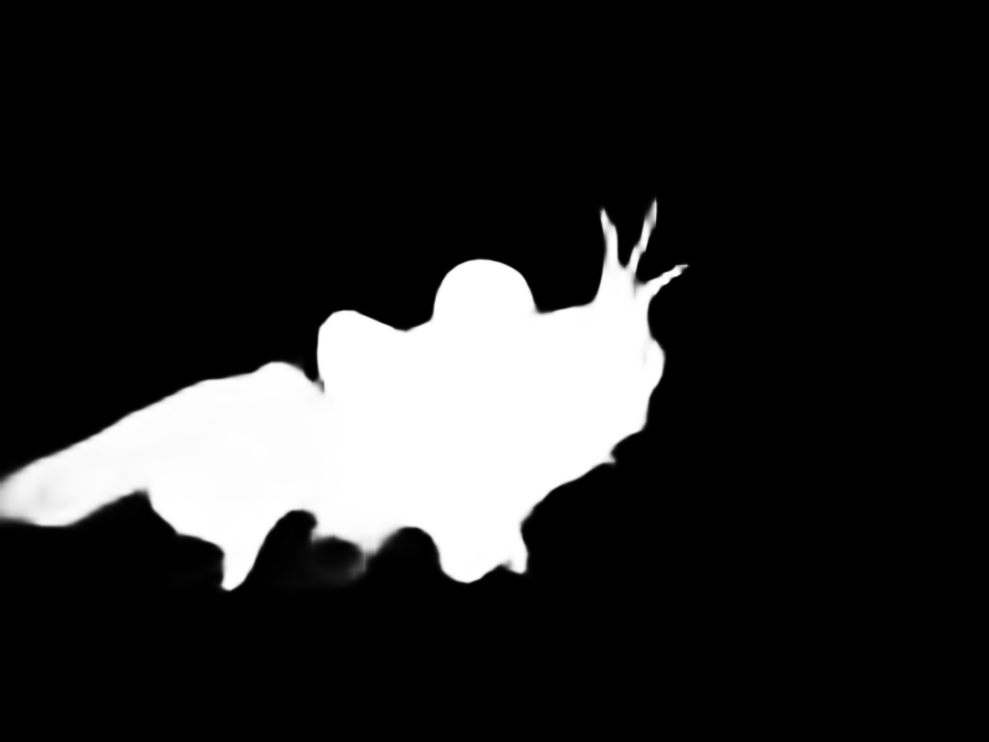}\vspace{-2pt}
%     \caption{+ CurriSeg}
	\end{subfigure} \\ \vspace{1mm}
\centering
		\begin{subfigure}{0.12\textwidth}
\setlength{\abovecaptionskip}{0.1cm}
		\centering 
\includegraphics[width=\textwidth]{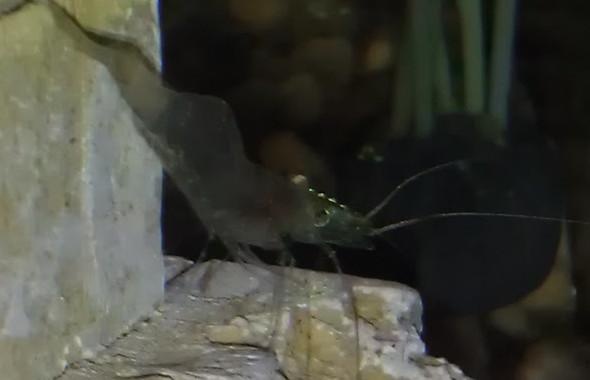}\vspace{-2pt}
		\caption{Origin}
	\end{subfigure}
	\begin{subfigure}{0.12\textwidth}  
\setlength{\abovecaptionskip}{0.1cm}
		\centering 
\includegraphics[width=\textwidth]{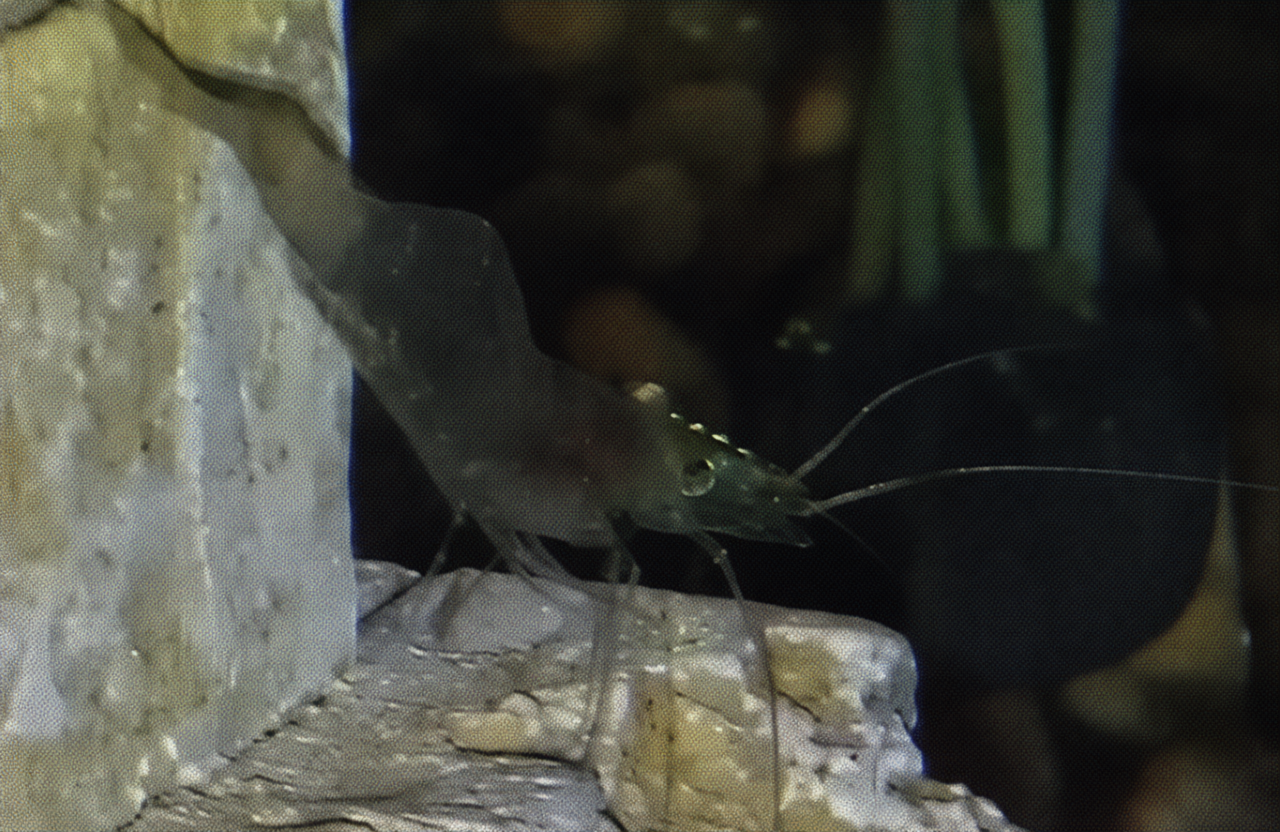}\vspace{-2pt}
		\caption{Degradation}
	\end{subfigure}
	\begin{subfigure}{0.12\textwidth}  
\setlength{\abovecaptionskip}{0.1cm}
		\centering 
\includegraphics[width=\textwidth]{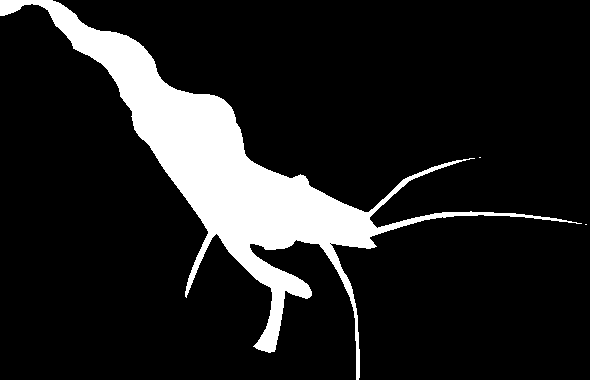}\vspace{-2pt}
\caption{GT}
	\end{subfigure}
    	\begin{subfigure}{0.12\textwidth} 
\setlength{\abovecaptionskip}{0.1cm}
		\centering 
\includegraphics[width=\textwidth]{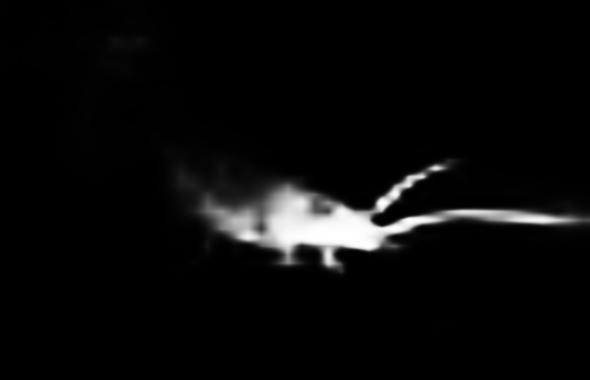}\vspace{-2pt}
\caption{Initial}
	\end{subfigure} 
		\begin{subfigure}{0.12\textwidth}
\setlength{\abovecaptionskip}{0.1cm}
		\centering 
\includegraphics[width=\textwidth]{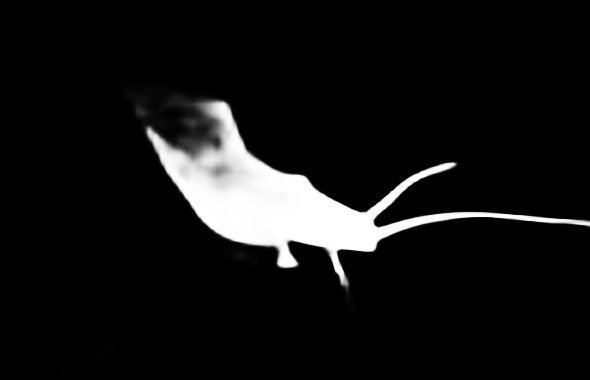}\vspace{-2pt}
		\caption{Mid}
	\end{subfigure}
	\begin{subfigure}{0.12\textwidth}  
\setlength{\abovecaptionskip}{0.1cm}
		\centering 
\includegraphics[width=\textwidth]{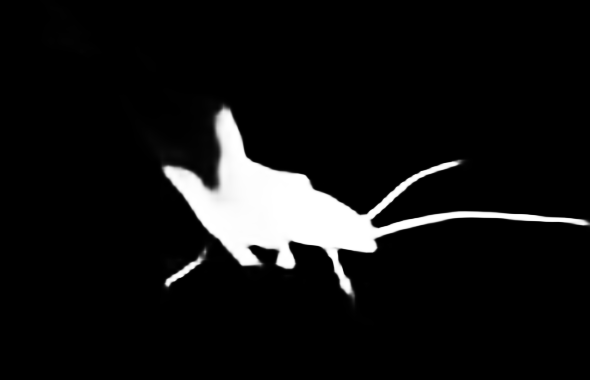}\vspace{-2pt}
		\caption{Final}
	\end{subfigure}
	\begin{subfigure}{0.12\textwidth}  
\setlength{\abovecaptionskip}{0.1cm}
		\centering 
\includegraphics[width=\textwidth]{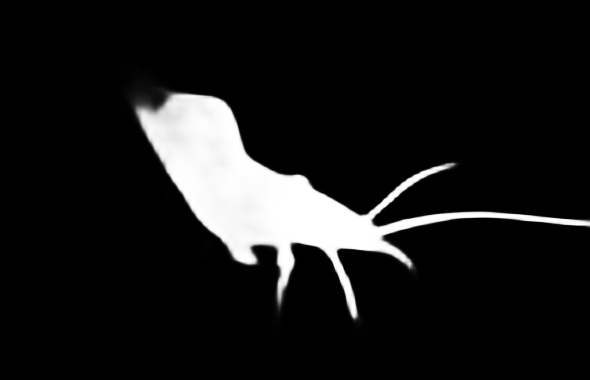}\vspace{-2pt}
\caption{+ Anti-Curri}
	\end{subfigure}
    	\begin{subfigure}{0.12\textwidth} 
\setlength{\abovecaptionskip}{0.1cm}
		\centering 
\includegraphics[width=\textwidth]{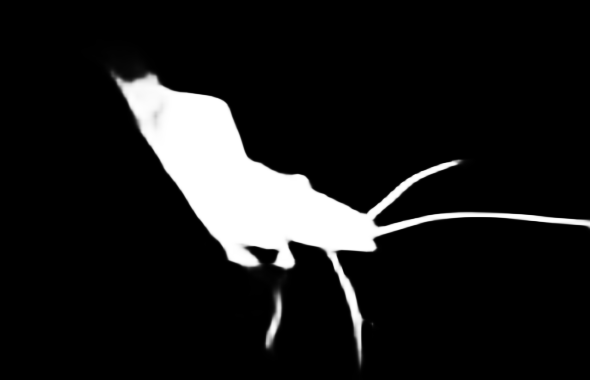}\vspace{-2pt}
\caption{+ CurriSeg}
	\end{subfigure}\vspace{-1mm}
\captionof{figure}{Robustness analysis under high-frequency loss (Row 1: haze) and interference (Row 2: noise). While converged FEDER (f) overfit fragile texture shortcuts, CurriSeg preserves structural integrity by balancing low-frequency context with task-relevant details.
% Qualitative analysis of frequency dependency. \textbf{Row 1}: High-frequency loss scenario (blur/haze). \textbf{Row 2}: High-frequency interference scenario (noise). Columns show: input, GT, baseline at mid-training, baseline at convergence, baseline + anti-curriculum only, and full CurriSeg. Mid-training models capture low-frequency structure but lack detail; converged baselines over-fit high-frequency cues and fail under distribution shift; anti-curriculum preserves structural robustness while CurriSeg achieves the best balance.
}
\label{fig:freq_insight}
\end{minipage}
\vspace{-5mm}
\end{table*}

\subsection{Ablation Study}\label{sec:ablation}
We conduct ablations on \textit{COD10K} using FEDER, a cutting-edge network that integrates frequency guidance and attention mechanisms, as the baseline in \cref{sec:ablation,sec:analysis}.

\noindent \textbf{Effect of robust curriculum selection}.
As shown in \cref{table:breakdown}, robust curriculum selection (WCS+PUE+TSSW) stabilizes training and suppresses ambiguous supervision. 
% is crucial for making curriculum learning beneficial in concealed scenes by stabilizing training and suppressing ambiguous supervision. 
In \cref{Table:AblationCurri}, replacing WCS with self‑paced learning~\cite{kumar2010self} or teacher-student curricula~\cite{matiisen2019teacher}  yields comparable results but adds complexity. 
Altering the warm-up schedule from $P(t)$ to $p_1(t)  = p_{min}+(1-p_{min})\cdot(\frac{t-1}{T_c-1})^2$ or $p_2(t)  = p_{min}+(1-p_{min})\cdot(\frac{t-1}{T_c-1})^{0.5}$ degrades performance, suggesting $p(t)$ offers a better stability-coverage trade-off. 
Within TSSW, dropping any component ($\omega_i^\mu$, $\omega_i^\sigma$, and $\omega_i^{out}$) or replacing $\omega_i^\sigma$ with $\omega_i^{\sigma1}=\text{max}(0, 1-\frac{|\tilde{\sigma}^2_i - \sigma^*|}{\gamma})$ and $\omega_i^{\sigma2}=1-(\frac{\tilde{\sigma}^2_i - \sigma^*}{\gamma})^2$ leads to noticeable drops. Finally, removing the entropy‑based pixel reweighting $\beta(t)$ or using exponential decay $\beta_1(t)=\text{exp}(-\frac{t}{T_c})$ also hurts.

\noindent \textbf{Effect of anti-curriculum promotion}.
As shown in \cref{Table:AblationAnti-Curri}, applying SBFT only to hard samples (top $50\%$) or to a random $50\%$ subset falls below CurriSeg. 
Replacing our circular low-pass mask with square or progressive filters worsens results.
% Changing the low‑pass mask from our circular filter to a square or progressive filter brings worse performance. 
with Gaussian blur, additive noise, texture-aware loss, or aggressive augmentation degrades metrics, confirming the benefit of principled spectral design.
% Replacing SBFT with generic spatial degradations, \textit{i.e.}, Gaussian blur or additive Gaussian noise, degrades all metrics, showing that SBFT benefits from a principled spectral design. 
Reversing the phase order causes dramatic performance drops.
% Likewise, using a texture‑aware loss or aggressive augmentation as the sole anti‑curriculum mechanism shows worse performance. Reverse CurriSeg swaps the order of curriculum and anti‑curriculum, leading to a dramatic performance drop. 

\noindent \textbf{Hyper-parameter sensitivity}.
\cref{fig:sensitivity} shows CurriSeg's sensitivity to key hyper-parameters. Optimal settings are: $K\!=\!10$, $p_{min}\!=\!0.6$ (WCS); $\sigma^*\!=\!0.5$, $\gamma\!=\!0.2$, $W_{min}^s\!=\!0.1$ (TSSW); $W_{min}\!=\!0.1$ (PUE); $r\!=\!0.95$, $T_c\!=\!60$ (ACP). The framework is robust within moderate ranges.
% We analyze the sensitivity of CurriSeg to key hyper-parameters in~\cref{fig:sensitivity}. In WCS, the optimal checkpoint interval is $K=10$, and the warm-up ratio peaks arounds $p_{min} = 0.6$. 
% For TSSW, CurriSeg is insensitive to $\sigma^*$ within a mid-range, while moderate $\gamma \approx 0.2 - 0.3$ and $ W_{min}^s \approx 0.1 - 0.2$ are preferred. For PUE, the best result occurs at $W_{min} \approx 0.1$. During the anti-curriculum phase, the SBFT cutoff ratio performs best around $r=0.95$, and the phase split  at $T_c=60$ is robust, favoring longer curriculum learning before SBFT.
% , consistent with the ``stabilize-then-perturb'' design.

\subsection{Further Analysis and Applications}\label{sec:analysis}

\noindent \textbf{Feature space analysis}. 
As shown in \cref{fig:tsne}, we visualize attention maps and feature distributions via t-SNE~\cite{maaten2008visualizing}. The baseline exhibits substantial cluster overlap, while CurriSeg training yields improved intra-class compactness and inter-class separation, confirming that our CurriSeg induces more discriminative representations.

% We visualize the attention map and feature distribution of foreground-background using t-SNE~\cite{maaten2008visualizing}. As shown in Fig.~\ref{fig:tsne}, the baseline exhibits substantial cluster overlap, reflecting CECS's inherent feature ambiguity. After CurriSeg training, the feature space demonstrates improved \textit{intra-class compactness} and \textit{inter-class separation}, confirming that our dual-phase curriculum induces more discriminative representations.
% To understand how CurriSeg shapes the learned representations, we visualize foreground and background feature distributions using t-SNE~\cite{maaten2008visualizing}. As illustrated in Fig.~\ref{fig:tsne}, the baseline model exhibits substantial overlap between foreground and background clusters, reflecting the inherent feature ambiguity in CECS tasks. After training with CurriSeg, the feature space shows markedly improved \textit{intra-class compactness} and \textit{inter-class separation}: foreground features form a tighter cluster while maintaining clear separation from background features. This visualization confirms that our dual-phase curriculum not only improves pixel-level predictions but also induces more discriminative intermediate representations, which is crucial for handling context-entangled scenarios where foreground and background share similar visual patterns.

\noindent \textbf{Broad generalization across settings and tasks}. 
We evaluate CurriSeg under weak/semi-supervision, multi-modal learning, and video segmentation (\cref{table:OtherSetting}), as well as semantic/instance segmentation, infrared target detection, and shadow detection (\cref{table:OtherTasks}). Consistent improvements demonstrate that our paradigm provides a general training principle for dense prediction, highlighting our potential.

\noindent \textbf{Performance dynamics across epochs}. 
\cref{fig:epoch_curve} shows that CurriSeg consistently surpasses baselines throughout training (averaged over five runs), with faster convergence and reduced variance indicating more stable optimization.
% Fig.~\ref{fig:epoch_curve} shows our performance (the average over five independent runs) across the training progress, where our CurriSeg always surpasses the baseline and helps the method coverage faster, validating our effectiveness. The reduced variance across runs further indicates more stable optimization.

\noindent \textbf{Robustness under image degradation}. 
We evaluate robustness by progressively introducing degraded samples (from NUN~\cite{he2025nested}) with ratios from 0\% to 100\%. As shown in \cref{fig:degradation}, CurriSeg outperforms baselines, with the gap widening at higher degradation levels, demonstrating enhanced reliance on low-frequency structural cues.

\noindent \textbf{Compatibility with advanced architectures}.
CurriSeg is architecture-agnostic. \cref{table:Compatibility} shows consistent gains on attention-based (FSEL), multi-scale (ZoomNet), uncertainty-aware (UGTR), and foundation model adapters (SAM series). ``+'': CurriSeg; ``*'': simple and specific adaptations (\cref{sec:CompatImplement}), highlighting broad applicability.

% \vspace{-2mm}
\section{Discussion}
% \vspace{-1mm}
CurriSeg offers a frequency-aware perspective extending beyond CECS. \textbf{(i) Spectral learning dynamics}: The ``stabilize-then-perturb'' trajectory first anchors the model on reliable representations through noise-resistant curriculum. \textbf{(ii) Mitigating high-frequency dependency}: SBFT's effectiveness suggests standard training drifts toward texture shortcuts. As shown in \cref{fig:freq_insight}, fully converged baselines often overfit fragile high-frequency cues, leading to collapse under spectral degradations. \textbf{(iii) Robustness via information bottlenecks}: By attenuating high-frequency components, CurriSeg enforces reliance on low-frequency boundaries and contextual semantics—a powerful, architecture-agnostic complement to frequency-aware modules.

% CurriSeg offers a frequency-aware perspective on dense prediction learning that extends beyond the CECS setting. \textbf{(i) Spectral learning dynamics}. By adopting a ``stabilize-then-perturb'' trajectory, CurriSeg first anchors the model on reliable representations through noise-resistant curriculum selection. \textbf{(ii) Mitigating High-Frequency Dependency}. The effectiveness of SBFT suggests that standard training typically drifts toward texture-based shortcuts—an inherent segmentation bias. As illustrated in \cref{fig:freq_insight}, while initial- and mid-training models capture robust low-frequency structures, fully converged baselines often overfit fragile high-frequency cues, leading to performance collapse under spectral degradations such as blur or noise. \textbf{(iii) Robustness via Information Bottlenecks}. By explicitly attenuating high-frequency components, CurriSeg creates an information bottleneck that enforces reliance on low-frequency boundaries and contextual semantics. These results demonstrate that late-stage regularization of texture shortcuts is a powerful, architecture-agnostic complement to modern frequency-aware or multi-scale modules, enhancing robustness without increasing inference overhead.

% \vspace{-2mm}
\section{Conclusions}
% \vspace{-1mm}
We proposed CurriSeg, which follows a ``stabilize-then-perturb'' process: robust curriculum selection uses temporal loss statistics and pixel-level uncertainty to construct a stable, noise-resistant learning schedule, and an anti-curriculum phase applies SBFT to attenuate high-frequency texture cues. This encourages reliance on structural and contextual information and reduces shortcut behavior.
Experiments show consistent performance and robustness gains without extra parameters or training overhead, highlighting our importance for entangled and dense visual scenes.

\section*{Acknowledgment}
This work was supported in part by the Foundation Fighting Blindness (BR-CL-0621-0812-DUKE and PPA-1224-0890-DUKE) and Research to Prevent Blindness Unrestricted Grant to Duke University. The first two authors contribute equally to this paper.

\section*{Impact Statement}
This work advances segmentation for concealed and context-entangled objects, with potential benefits in medical image analysis and autonomous perception. As with other segmentation systems, deployment in safety-critical settings such as clinical diagnosis or driving requires careful validation, since segmentation errors may carry real-world consequences. We do not foresee additional ethical concerns beyond those common to general dense-prediction research.

\bibliography{example_paper}
\bibliographystyle{icml2026}

%%%%%%%%%%%%%%%%%%%%%%%%%%%%%%%%%%%%%%%%%%%%%%%%%%%%%%%%%%%%%%%%%%%%%%%%%%%%%%%
%%%%%%%%%%%%%%%%%%%%%%%%%%%%%%%%%%%%%%%%%%%%%%%%%%%%%%%%%%%%%%%%%%%%%%%%%%%%%%%
% APPENDIX
%%%%%%%%%%%%%%%%%%%%%%%%%%%%%%%%%%%%%%%%%%%%%%%%%%%%%%%%%%%%%%%%%%%%%%%%%%%%%%%
%%%%%%%%%%%%%%%%%%%%%%%%%%%%%%%%%%%%%%%%%%%%%%%%%%%%%%%%%%%%%%%%%%%%%%%%%%%%%%%
\newpage
\appendix
\onecolumn
\setcounter{figure}{0}
\renewcommand{\figurename}{Fig.}
\renewcommand{\thefigure}{S\arabic{figure}}
\setcounter{table}{0}
\renewcommand{\tablename}{Table}
\renewcommand{\thetable}{S\arabic{table}}

\end{document}